\documentclass[10pt,twocolumn,letterpaper]{article}

\usepackage[pagenumbers]{cvpr} 

\usepackage{algorithm}
\usepackage{algorithmic}
\usepackage{amsmath}
\usepackage{amsfonts}
\usepackage{amssymb}
\usepackage{booktabs}
\usepackage{graphicx}
\usepackage{mathrsfs}
\usepackage{multirow}
\usepackage{xcolor}

\DeclareMathOperator{\vect}{vec}

\DeclareMathOperator{\cat}{Cat}
\DeclareMathAlphabet{\mymathbb}{U}{BOONDOX-ds}{m}{n}

\definecolor{cvprblue}{rgb}{0.21,0.49,0.74}
\usepackage[pagebackref,breaklinks,colorlinks,allcolors=cvprblue]{hyperref}

\newcounter{eqfn} 
\makeatletter
\def\equalcontrib{%
    \ifnum\value{eqfn}=0\relax
        \thanks{These authors contributed equally.}%
        \setcounter{eqfn}{\value{footnote}}%
    \else
        \footnotemark[\arabic{eqfn}]%
    \fi
}
\makeatother

\title{Unsupervised Training of Diffusion Models for Feasible Solution Generation in Neural Combinatorial Optimization}

\author{
Seong-Hyun Hong\textsuperscript{\rm 1}\equalcontrib, \
Hyun-Sung Kim\textsuperscript{\rm 1}\equalcontrib, \
Zian Jang\textsuperscript{\rm 1},
Deunsol Yoon\textsuperscript{\rm 2},
Hyungseok Song\textsuperscript{\rm 2},
Byung-Jun Lee\textsuperscript{\rm 1, 3}\\
\textsuperscript{\rm 1}Korea University,
\textsuperscript{\rm 2}LG AI Research,
\textsuperscript{\rm 3}Gauss Labs Inc.\\
Seoul, Republic of Korea\\
}

\begin{document}
\maketitle
\begin{abstract}
Recent advancements in neural combinatorial optimization (NCO) methods have shown promising results in generating near-optimal solutions without the need for expert-crafted heuristics. However, high performance of these approaches often rely on problem-specific human-expertise-based search after generating candidate solutions, limiting their applicability to commonly solved CO problems such as Traveling Salesman Problem (TSP).
In this paper, we present IC/DC, an unsupervised CO framework that directly trains a diffusion model from scratch. We train our model in a self-supervised way to minimize the cost of the solution while adhering to the problem-specific constraints. 
IC/DC is specialized in addressing CO problems involving two distinct sets of items, and it does not need problem-specific search processes to generate valid solutions. 
IC/DC employs a novel architecture capable of capturing the intricate relationships between items, and thereby enabling effective optimization in challenging CO scenarios. IC/DC achieves state-of-the-art performance relative to existing NCO methods on the Parallel Machine Scheduling Problem (PMSP) and Asymmetric Traveling Salesman Problem (ATSP).
\end{abstract}    
\vspace{-10pt}
\section{Introduction}
\label{sec:intro}

Combinatorial optimization (CO) aims to find the optimal solution that maximizes or minimizes an objective function from a large, discrete set of feasible solutions. This field has been extensively studied due to its broad industrial applications, including logistics, supply chain optimization, job allocation, and more~\citep{zhang2023review}. Despite its significance, many CO problems are NP-complete, and developing efficient approximation algorithms is essential.

Traditionally, approximation algorithms for CO have been developed using mathematical programming or hand-crafted heuristics~\citep{miller1960integer, lkh3}. 
However, the need for problem-specific expertise and the high computational demands of these methods has sparked increasing interest in applying {Neural Combinatorial Optimization (NCO)}, deep learning techniques to CO problems. Early deep learning approaches framed CO problems as sequential decision-making tasks, generating solutions in an \textit{autoregressive manner}~\citep{bello2016neural, kool2018attention}.
However, these methods were relatively limited in performance due to their inability to revise previously made decisions. 

In contrast, \textit{heatmap-based methods} construct an initial (possibly infeasible) solution, i.e. heatmap, and iteratively refine the heatmap through corrections and adjustments~\citep{sun2023difusco, min2024unsupervised}. 
By allowing for the revision of earlier decisions, heatmap-based methods overcome the limitations of autoregressive approaches and avoid the compounding errors typically associated with early decisions.

\begin{figure*}[t]
\centering
\includegraphics[width=0.92\textwidth]{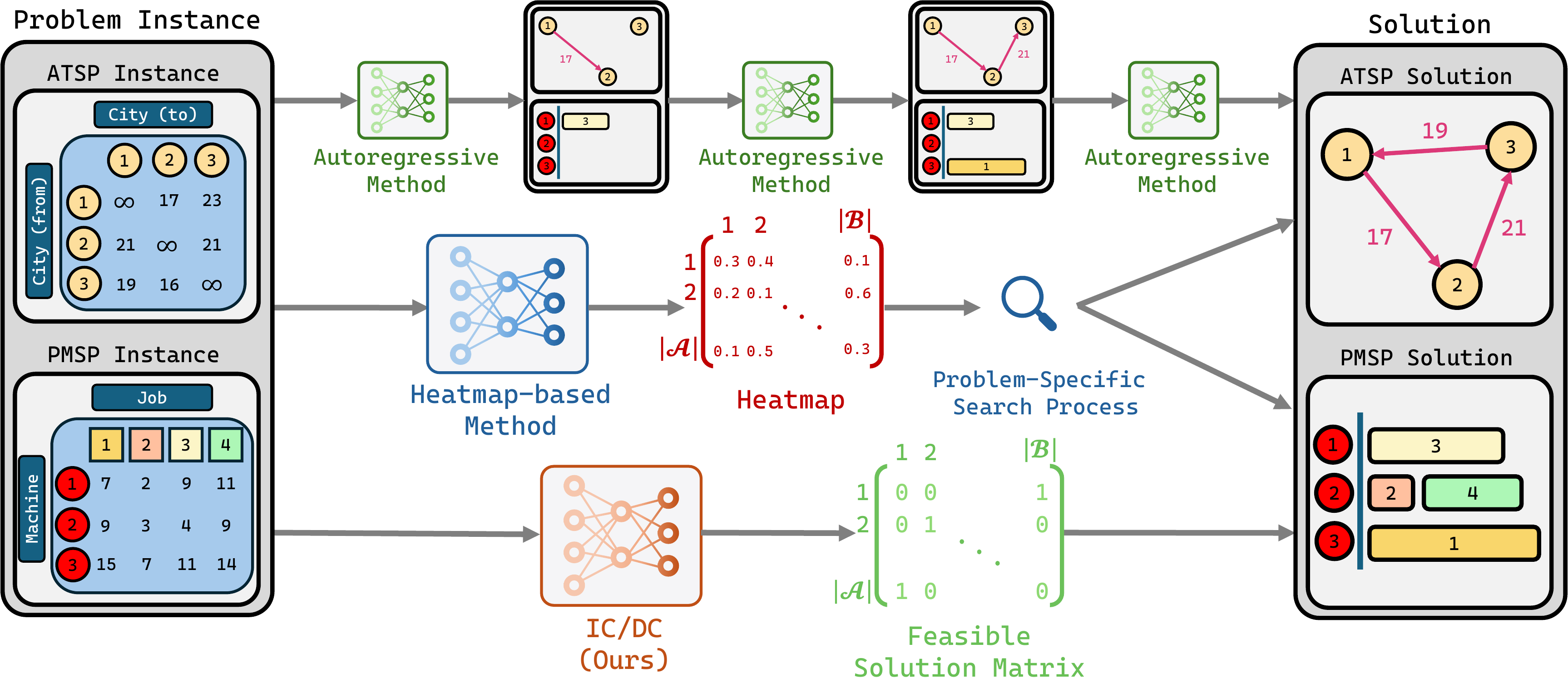} 
\caption{ 
The figure illustrates the CO problems we aim to address, and compares various learning-based approaches for CO problems. \textbf{(Left)} We focus on problems involving up to two distinct sets of items, which can be represented as a matrix. In ATSP, rows and columns represent cities (from/to), with each matrix element indicating distance. In PMSP, rows and columns represent machines and jobs, with each element representing processing time. \textbf{(Center)} The diagram highlights how these three approaches differ in generating solutions. Autoregressive methods generate solutions by iterating through items, while heatmap-based methods create a heatmap followed by a problem-specific heatmap search process. Our proposed IC/DC approach generates feasible solutions directly, using a training procedure that guides the diffusion model to satisfy the constraints.
}
\label{fig1}
\end{figure*}

Despite the impressive performance of heatmap-based methods, previously proposed algorithms had several significant drawbacks, such as the need for costly supervision~\citep{sun2023difusco}, or use of problem-specific objective that aren't applicable to other CO problems~\citep{min2024unsupervised}. More importantly, due to the challenges in imposing constraints on heatmap generation, previous studies relied on problem-specific search process to extract feasible solutions from possibly infeasible heatmaps. Designing these problem-specific search requires specialized knowledge and cannot be easily adapted to different CO problems.

In this work, we propose a novel method for training a feasible solution generator for NCO with a diffusion model in a self-supervised manner, eliminating the need for costly supervision, problem-specific objectives, or problem-specific search process. 
We demonstrate our approach on two distinct and challenging CO problems: the parallel machine scheduling problem (PMSP) and the asymmetric traveling salesman problem (ATSP), each involving two different classes of items where item features are defined by an asymmetric matrix representing their interrelationships~\citep{kwon2021matrix}. 
We show that our method achieves state-of-the-art performance among deep learning approaches.

In summary, the main contributions of this paper are:
\begin{itemize}
    \item To the best of our knowledge, this is the first study to introduce a diffusion model for combinatorial optimization (CO) involving two sets of items.
    
    \item We propose \textbf{IC/DC}, a novel method for training a diffusion model in a self-supervised manner, eliminating the need for costly supervision, problem-specific objectives, or problem-specific search process.
    
    \item We empirically demonstrate that our model improves upon other learning-based methods across two distinct CO problems.
\end{itemize}

\vspace{-10pt}
\section{Related Works}
\label{sec:related_works}

For widely-studied CO problems like the travelling salesman problem (TSP), several off-the-shelf solvers are available, such as CPLEX~\citep{cplex} and OR-tools~\citep{ortools}. These solvers are build on a variety of heuristics, incorporating search methods~\citep{lkh3}, mathematical programming~\citep{arora1996polynomial}, and graph algorithms~\citep{christofides2022worst}. Typically, these approaches rely on problem-specific, handcrafted techniques, which limits their flexibility in adapting to diverse variants encountered in real-world scenarios. 
\vspace{-10pt}
\paragraph{Autoregressive methods} To overcome this limitation, learning-based solvers have been developed, with early studies primarily focusing on autoregressive approaches. \citet{bello2016neural} were the first to propose solving CO problems using a pointer network trained via reinforcement learning (RL). \citet{kwon2021matrix} introduced an architecture called MatNet, which builds on the graph attention network (GAT) to encode various types of objects, enabling it to tackle more complex CO problems. While these autoregressive models offer fast solution generation and can manage intricate CO problems, they are limited by their inability to revise previously made decisions. Recent advancements have addressed these limitations by incorporating iterative refinement techniques, mainly utilizing heatmap-based methods and outperformed autoregressive methods in terms of performance.
\vspace{-15pt}
\paragraph{Heatmap-based methods}

To the best of our knowledge, two notable works have effectively addressed CO problems by applying heatmap-based approaches~\citep{sun2023difusco, min2024unsupervised}. Both were able to achieve solution quality comparable to off-the-shelf solvers while significantly reducing generation time. DIFUSCO~\citep{sun2023difusco} is a graph neural network (GNN)-based diffusion model trained in a supervised manner to replicate solutions generated by solvers. Similarly, UTSP~\citep{min2024unsupervised} employs a GNN-based model trained in an unsupervised manner, eliminating the need for costly solution generation from traditional solvers. However, UTSP's objective is based on the concept of the Hamiltonian cycle, limiting its applicability to TSP. While CO problems typically impose strict constraints on solutions, such as forming a Hamiltonian cycle, these algorithms are not trained to inherently satisfy those constraints. Instead, they rely on additional search techniques, such as active search methods~\citep{qiu2022dimes} or Monte Carlo Tree Search (MCTS)~\citep{silver2016mastering,fu2021generalize}, to produce feasible solutions. This reliance limits their applicability to CO problems with varying constraints on solutions.
\section{Improving Combinatorial Optimization through Diffusion with Constraints}
\label{sec:improving_co_through_diffusion_w_constraints}

To combine the high-quality solutions of heatmap-based methods with the flexibility of autoregressive approaches, we propose \textbf{I}mproving \textbf{C}ombinatorial optimization through \textbf{D}iffusion with \textbf{C}onstraints (IC/DC). This approach ensures the feasibility of solutions while training diffusion model in a self-supervised manner, eliminating the need for costly supervision and problem-specific search processes.

\subsection{Problem Definition}

We consider a family of CO problems $\mathcal{C}$, which involve two distinct sets of items. Each problem $c\in \mathcal{C}$ is defined by two sets of items $\mathcal{A}$ and $\mathcal{B}$, and matrices that describe the relationships between these two sets of items, as illustrated on the left side of~\cref{fig1}.
The solution to a CO problem $c$ is represented by a binary matrix \(X\in \mathcal{X}=\{0, 1\}^{|\mathcal{A}| \times |\mathcal{B}|}\). 

Typically, for each problem $c$, there exists a feasible set of solutions, and a particular solution $X$ is evaluated using a problem-specific scoring function $\mathrm{score}:\mathcal{X}\times\mathcal{C}\rightarrow\mathbb{R}$ when it is feasible. 
For clarity, we define the reward function \(R: \mathcal{X} \times \mathcal{C} \rightarrow \mathbb{R}\) as follows:
\begin{equation}
R(X,c) = 
\begin{cases}
    \mathrm{score}(X,c) & \text{if $X$ is feasible for $c$} \\
     -\infty & \text{otherwise}
\end{cases},
\end{equation}
which allows us to express the objective of the CO problem $c$ as $\max_{X \in \mathcal{X}}\, R(X, c)$.

\subsection{Training of IC/DC}
We build upon a discrete diffusion model with categorical corruption processes~\citep{austin2021structured}. We represent the uncorrupted solution that we aim to generate as $X_0$, with the corrupted latent variables denoted as $X_1,...,X_T$. We use lowercase $x$ to represent the vectorized forms of $X$, where $x_t=\vect(X_t)\in \{0, 1\}^{|\mathcal{A}||\mathcal{B}|}$, and tilded $\tilde{x}$ to denote the one-hot encoded versions, $\tilde{x}_t\in \{0, 1\}^{|\mathcal{A}||\mathcal{B}|\times 2}$. In line with diffusion model conventions, we use $q(\cdot)$ to denote the data distribution/generative forward process, while $p_\theta(\cdot)$ represents the denoising reverse process, which is learned to generate the solutions.
\vspace{-5pt}
\paragraph{Forward process} 
Our forward process is defined as:

\begin{align}\label{forward_process_eq}
q(X_t|X_{t-1})&= \cat(\vect^{-1}(\tilde{x}_{t-1}Q_{t})), \\
q(X_{t}|X_0)&= \cat(\vect^{-1}(\tilde{x}{Q}_{1:t})), \\
q(X_{t-1}|X_t, X_0) &=  \cat\left(\vect^{-1}\left(\frac{\tilde{x}_t Q_t^\top \odot \tilde{x}_0 Q_{1:t-1}}{\tilde{x}_0 Q_{1:t} \tilde{x}_t^\top}\right)\right),
\end{align}

where \(Q_{1:t} = Q_{1}Q_2...Q_{t} \), $\odot$ denotes the element-wise multiplication, and $\vect^{-1}$ reshapes the input to the shape $|\mathcal{A}|\times |\mathcal{B}|\times 2$. The matrix $Q_t\in [0,1]^{2 \times 2}$ is a noise transition matrix that independently applies noise to each element of the solution matrix.
We design this noise transition matrix to align with the prior distribution of feasible solutions $\bar{q}$ in the limit~\citep{vignac2022digress}, such that $\lim_{T\rightarrow \infty} Q_{1:T}z=\bar{q}$ for any vector $z$. This is achieved by defining:
\begin{equation}
    Q_t=\alpha_t I+\beta_t\mymathbb{1}\bar{q}^\top,
\end{equation}
where $\mymathbb{1}$ is vector of ones, and $\alpha_t$ and $\beta_t$ are scheduled appropriately with typical diffusion schedulers. The formulas for computing $\bar{q}$ for the CO problems demonstrated in the experiments are detailed in~\cref{qbar_formula}.
\vspace{-5pt}
\paragraph{Reverse process} We follow the parametrization of~\citet{austin2021structured}, where neural network $f_\theta(\cdot)$ is trained to directly predict logits of $X_0$ from each $X_t$, as follows:
\begin{align}
    p_\theta(X_0|X_t,c)&=\cat(f_\theta(X_t, t,c)), \\
    p_{\theta}(X_{t-1}|X_t, c) &\propto \sum_{X_0} q(X_{t-1},X_t| X_0)p_{\theta}(X_{0}|X_t, c).
    \label{reverse_process_eq}
\end{align}

\begin{algorithm}[tb]
\caption{Training IC/DC}
\label{icdc algorithm}
\textbf{Input}: A set of CO problems $\mathcal{C}$, learning late $\gamma$, diffusion step $T$, target mix ratio $\alpha$
\begin{algorithmic}[1] 
\STATE \textbf{function} \texttt{CLONING}(${\theta}$)
\begin{ALC@g}
    \STATE $X_0, c \sim \mathcal{D}_{\tilde{q}}(\alpha)$ with a probability proportional to $\exp(R(X_0,c))$
    \STATE $t \sim \text{Uniform}(\{1, ..., T\})$
    \STATE $X_t \sim q(X_t|X_0)$ 
    \STATE $\theta\gets\theta - \gamma\nabla_{\theta}\mathcal{L}_{\texttt{CLN}}$ according to~\cref{eq:cloning} 
\end{ALC@g}
\STATE \textbf{end function}
\item[]
\STATE \textbf{function} \texttt{IMPROVEMENT}(${\theta}$)
\begin{ALC@g}
    \FOR{$i \gets 1$ to $N$}
    \STATE Sample $c^{(i)}$ from a set of CO problems $\mathcal{C}$
    \STATE $X_T^{(i)} \sim q(X_T^{(i)})$
    \FOR{$t\gets T$ to $1$}
        \STATE ${X}_{t-1}^{(i)} \sim p_{\theta}(X_{t-1}|{X}_{t}^{(i)},c^{(i)})$
    \ENDFOR
    \STATE $\widehat{X}_0^{(i)} \sim 
    \widehat{p}_\theta(X_0|X_1^{(i)}, c^{(i)})$
    \ENDFOR
    \STATE $\theta \gets \theta - \gamma\nabla_{\theta}\mathcal{L}_{\texttt{IMP}}(\theta)$ with $\{(\widehat{X}_0^{(i)},c^{(i)})\}_{i=1}^N$ according to~\cref{improve term}
    \STATE Sample $\{(\tilde{X}_0^{(i)}, \tilde{c}^{(i)})\}_{i=1}^{\frac{1-\alpha}{\alpha}N}$ from $q(X_0)$, $ \mathcal{C}$
    \STATE Store $\{(X_0^{(i)},c^{(i)})\}_{i=1}^N,\{(\tilde{X}_0^{(i)}, \tilde{c}^{(i)})\}_{i=1}^{\frac{1-\alpha}{\alpha}N}$ in $\mathcal{D}_{\tilde{q}}$
\end{ALC@g} 
\STATE \textbf{end function}
\item[]
\STATE Initialize an empty set $\mathcal{D}_{\tilde{q}}=\{\}$
\REPEAT
    \FOR{ $m\leftarrow 1$ to $M$}
    \STATE \texttt{CLONING}(${\theta}$, $\mathcal{D}_{\tilde{q}}$)
    \ENDFOR
    \STATE \texttt{IMPROVEMENT}(${\theta}$, $\mathcal{D}_{\tilde{q}}$)
\UNTIL convergence
\end{algorithmic}
\end{algorithm}

Although this parameterization facilitates the easy computation of diffusion loss when a target dataset is provided, samples from $p_\theta$ may not satisfy the constraints since each element of the solution matrix is independently sampled from a Categorical distribution. This limitation required the use of a feasibility-enforcing search process in previous heatmap-based studies on CO~\citep{sun2023difusco,min2024unsupervised}. However, this approach requires a search process specifically tailored to each CO problem, and there is no assurance that the search process will preserve the quality of the solution that the diffusion model aims to generate.
\vspace{-5pt}
\paragraph{Feasibility-enforced generation} To ensure the generation of feasible solutions, we design a process inspired by autoregressive methods, which we call the feasibility-enforced generation process. In this approach, we sample one element of the solution matrix at a time, ensuring its feasibility based on the previously sampled elements, as follows:
\begin{align}
    \widehat{p}_\theta(X_0|X_t, c)&\propto\prod_{i=1}^{|\mathcal{A}||\mathcal{B}|}\cat([X_0]_i|[f_\theta(X_t, t, c)]_i)\mathbb{I}([X_0]_i),\\
    \mathbb{I}([X_0]_i)&=\begin{cases}
        1 & \text{$[X_0]_i$ is feasible given $[X_0]_{1:i-1}$}\\
        0 & \text{otherwise}
    \end{cases}.
\end{align}
This approach, similar to the flexibility of autoregressive methods, allows for the straightforward enforcement of feasibility in the generated samples.
However, $\widehat{p}_\theta$ cannot be directly utilized as the reverse process because it involves discrete sampling, which prevents the use of the reparameterization trick, thereby making conventional and efficient variational training methods inapplicable. When using $p_\theta$ as the reverse process, the connection between $p_\theta$ and $\widehat{p}_\theta$ becomes weak, leading to a lack of guarantee that samples from $\widehat{p}_\theta$ will retain the desired characteristics.
\paragraph{Alternating training} To this end, we propose an iterative training approach that alternates between the \texttt{CLONING} step and the \texttt{IMPROVEMENT} step. In the \texttt{CLONING} step, we update the reverse process $p_\theta$ by maximizing the (lower bound of the) log likelihood of a set of high scoring feasible solutions: the surrogate targets. This guides $p_\theta$ toward generating high-quality feasible solutions, and strengthen its alignment with $\widehat{p}_\theta$, as they become identical when $p_\theta$ generates feasible samples only. In the \texttt{IMPROVEMENT} step, we directly update $\widehat{p}_\theta$ using reinforcement learning to maximize the scores of generated solutions. These two steps work in tandem, the \texttt{IMPROVEMENT} step is direct but computationally intensive, and \texttt{CLONING} step is more efficient but is only an indirect method of improving samples from $\widehat{p}_\theta$.
\paragraph{Surrogate targets}
In standard diffusion model training, the target distribution $q(X_0|c)$, which represents the distribution of optimal solutions given a problem $c$, is typically available.
However, in CO problems, obtaining such supervised dataset is often prohibitively expensive. To address this, we propose training our diffusion model in a self-supervised manner using a surrogate target distribution \(\tilde{q}(X_0|c)\) instead. 

This surrogate distribution is progressively refined during training and is defined as a reward-weighted mixture of two distributions:
\begin{align}
\tilde{q}(X_0|c) &\propto \exp(R(X_0,c))[(1-\alpha)q(X_0) + \alpha p_{\theta}(X_{0}|c)],
\end{align}
where the mixture is controlled by the hyperparameter $\alpha\in[0,1]$. 
In the initial stage of training, the solutions generated by reverse process $p_\theta(X_0|c)$ are not feasible in general. This leads to using more samples from the prior distribution of feasible solutions $q(X_0)$, and guide the diffusion model to generate feasible solutions. As training progresses and as $p_{\theta}(X_{0}|c)$ begins to generate feasible solutions, the reward-weighting allows the diffusion model to refine itself by focusing on its high-scoring, feasible generations. Meanwhile, the inclusion of prior distribution of feasible solutions \(q(X_0)\) introduces diversity to the training, counteracting the tendency of $p_{\theta}$ to become too narrow as training progresses.

\begin{figure*}[t]
\centering
\includegraphics[width=0.8\textwidth]{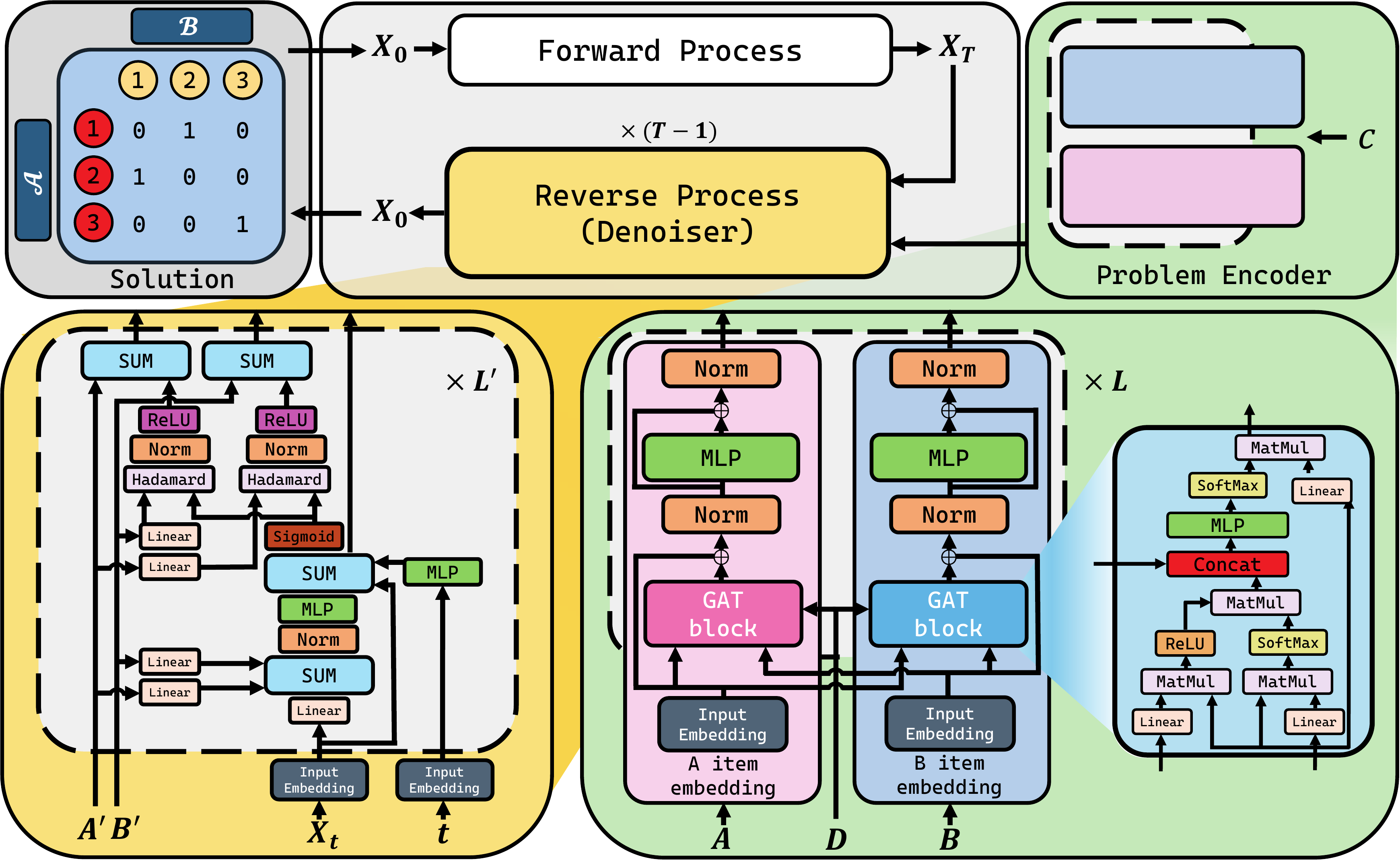} 
\caption{This figure illustrates our architecture and diffusion process. \textbf{(Bottom-right)} The problem encoder encodes a problem instance $c=(A, B, D)$ into a problem embedding $(A',B')$. \textbf{(Bottom-left)} The denoiser takes a problem embedding, noisy solution and a timestep embedding $(A', B', X_t,t)$ and outputs denoised solution $X_0$.}\label{fig:architecture}
\end{figure*}

\paragraph{\texttt{CLONING} step}
With the surrogate target distribution defined above, we perform standard diffusion training by minimizing the KL-divergence between the model's generative distribution and the surrogate target distribution:
\begin{equation}\label{minimize_kl}
    \min_{\theta}D_{KL}(\tilde{q}(X_0|c) \parallel p_\theta(X_0|c)),
\end{equation}
which results in a variational bound objective (see~\cref{reward weighed diffusion loss} for detailed derivations), $\mathcal{L}_{\text{VB}}(\theta):=$
\begin{equation}
    \mathbb{E}_{X_0\sim\tilde{q}} 
    \left[\sum_{t=2}^{T} D_{KL}(q(X_{t-1}|X_{t}, X_0) \| p_\theta(X_{t-1}|X_t, c))\right]
\end{equation}

Inspired by recent practices~\citep{austin2021structured}, we also incorporate the following auxiliary losses:
\begin{align}
\mathcal{L}_{\text{prd}}(\theta)&:=\mathbb{E}_{X_0\sim \tilde{q}(X_0|c), X_t\sim q(X_t|X_0)}\left[-\log p_\theta(X_0|X_t, c)\right],\\
\mathcal{L}_{\text{cst}}(\theta)&:=\mathbb{E}_{X_0\sim \tilde{q}(X_0|c), X_t\sim q(X_t|X_0)}\left[C( p_\theta(X_0|X_t, c))\right],
\end{align}
where $\mathcal{L}_{\text{prd}}$ encourages accurate predictions of the data $X_0$ at each time step, and $\mathcal{L}_{\text{cst}}$ discourages infeasible predictions, with $C(\cdot)$ being a differentiable function that approximately measures constraint violations of samples using the Gumbel-softmax trick (see~\cref{constraint loss} for details on $C$).

In summary, during the \texttt{CLONING} step, we minimize:
\begin{equation}
\mathcal{L}_{\texttt{CLN}}(\theta)=\mathcal{L}_{\text{VB}}(\theta)+\lambda_1 \mathcal{L}_{\text{prd}}(\theta) + \lambda_2 \mathcal{L}_{\text{cst}}(\theta).
\label{eq:cloning}
\end{equation}

\paragraph{\texttt{IMPROVEMENT} step}

To directly improve the feasibility-enforced generations, we minimize the following objective:
\begin{equation}
    \mathcal{L}_{\texttt{IMP}}(\theta)=-\mathbb{E}_{X_0\sim \widehat{p}_\theta(X_0|c)}[R(X_0,c)].
\end{equation}
Similar to autoregressive methods, the feasibility-enforced generation process can be viewed as a sequential decision making task, where each element of the solution matrix $X_0$ is determined step by step. This perspective allows us to compute the gradient of the above objective using the REINFORCE algorithm~\citep{williams1992simple}. 

After sampling a set of solutions $\{X_0^{(1)}, X_0^{(2)}, ..., X_0^{(N)}\}$ using $\widehat{p}_\theta(X_0|c)$, we approximate the gradient as follows: $\nabla_{\theta} \mathcal{L}_{\texttt{IMP}}(\theta) \approx$
\begin{equation}\label{improve term}
 -\frac{1}{N}\sum_{i=1}^N\left(R^{(i)}-\frac{1}{N}\sum_{j=1}^NR^{(j)}\right) \nabla_{\theta} \log \widehat{p}_\theta\left(X_0^{(i)}|c\right),
\end{equation}
where $R^{(i)}=R(X_0^{(i)}, c)$. This approach is adapted from the baseline estimation method of POMO~\citep{kwon2020pomo}.

\paragraph{Summary} We train IC/DC by alternating between the \texttt{CLONING} step--diffusion model training with surrogate targets--and the \texttt{IMPROVEMENT} step--reinforcement learning of feasibility-enforced generation, as shown in~\cref{icdc algorithm}. In practice, we use a replay memory $\mathcal{D}_{\tilde{q}}$ to implement a surrogate target distribution, and perform multiple \texttt{CLONING} steps for each \texttt{IMPROVEMENT} step. This is because \texttt{CLONING} updates only a single timestep of the diffusion model, and \texttt{IMPROVEMENT} is significantly slower due to the online generations.
\section{Architecture}
\label{sec:architecture}

The neural network we need is $f:\mathcal{X}\times\mathcal{C}\rightarrow\mathcal{X}$, which encodes the CO problem and outputs a distribution over binary matrices given an input binary matrix. To achieve this, we propose a \textit{problem encoder} that effectively encodes CO problem, and a \textit{denoiser}, a specialized variant of GNN that processes the bipartite graph between two sets of items.
\paragraph{Problem encoder} For the CO problems we consider, a problem instance $c$ consists of information about two sets of items and their relationships. For simplicity, let's assume that all items share the same number of features $d$; if not, different embedding layers can be used to standardize the feature dimensions. 
The information for the items in set $\mathcal{A}$ is represented by the matrix $A\in \mathbb{R}^{|\mathcal{A}|\times d}$, and similarly, the items in set $\mathcal{B}$ are represented by the matrix $B\in \mathbb{R}^{|\mathcal{B}|\times d}$. There relationship between these items is captured by the matrix $D\in\mathbb{R}^{|\mathcal{A}|\times |\mathcal{B}|}$. Together, these matrices define the problem instance, i.e., $c=(A, B, D)$.

To effectively encode the problem represented by these matrices, we adopt the dual graph attentional layer structure from MatNet~\citep{kwon2021matrix}, but replace the attention layer with a modified version of graph attention networks (GAT, \citep{velivckovic2017graph} 2017) that is specifically designed to process a bipartite graph. The problem encoder consists of $L$ layers, where each layer takes $(A, B, D)$ as input and outputs updated features $(A', B')$. The outputs of the final layer are then passed to the denoiser. The bottom-right side of~\cref{fig:architecture} illustrates this problem encoder. Detailed equations of problem encoder layer are provided in~\cref{sec:problem_encoder}.

\paragraph{Denoiser} Building on recent empirical successes~\citep{joshi2020learning, qiu2022dimes}, we extend the anisotropic graph neural network (AGNN) to handle bipartite graphs, allowing us to consider two distinct sets of items, and use it as the denoiser. The input embedding layer maps each element of the noisy solution matrix $X_t$ and the timestep $t$ into $d$-dimensional features. These embeddings are then passed to the AGNN, along with the problem embedding $A'$ and $B'$. After $L'$ layers of AGNN, the embedded solution matrix with updated features is passed through a linear layer to produce the denoised solution matrix $X_0$.
The bottom-left side of~\cref{fig:architecture} illustrates this denoiser. 
Detailed equations of denoiser layer are provided in~\cref{sec:denoiser}.
\section{Experiments}
\label{sec:experiments}

\begin{table*}[h]
    \centering
    \setlength{\tabcolsep}{1.1pt} 
    \renewcommand{\arraystretch}{1.0} 
    \scriptsize 
    \begin{tabular}{llcccccccccccccccccc}
        \toprule
        & & \multicolumn{9}{c}{PMSP-20} & \multicolumn{9}{c}{PMSP-50} \\
        \cmidrule(lr){3-11} \cmidrule(lr){12-20}
        & & \multicolumn{3}{c}{3 $\times$ 20} & \multicolumn{3}{c}{4 $\times$ 20} & \multicolumn{3}{c}{5 $\times$ 20} 
          & \multicolumn{3}{c}{3 $\times$ 50} & \multicolumn{3}{c}{4 $\times$ 50} & \multicolumn{3}{c}{5 $\times$ 50} \\
        \cmidrule(lr){3-5} \cmidrule(lr){6-8} \cmidrule(lr){9-11} \cmidrule(lr){12-14} \cmidrule(lr){15-17} \cmidrule(lr){18-20}
        & Method & M.S. $\downarrow$ & Gap $\downarrow$ & Time $\downarrow$ 
                 & M.S. $\downarrow$ & Gap $\downarrow$ & Time $\downarrow$ 
                 & M.S. $\downarrow$ & Gap $\downarrow$ & Time $\downarrow$
                 & M.S. $\downarrow$ & Gap $\downarrow$ & Time $\downarrow$
                 & M.S. $\downarrow$ & Gap $\downarrow$ & Time $\downarrow$
                 & M.S. $\downarrow$ & Gap $\downarrow$ & Time $\downarrow$ \\
        \midrule
        Baseline & CP-SAT 
        & 42.63 & 0\% & (1m) 
        & 28.11 & 0\% & (1m) 
        & 20.58 & 0\% & (3m) 
                                     & 102.06 & 0\% & (2m) 
                                     & 65.90 & 0\% & (3m) 
                                     & 47.03 & 0\% & (5m) \\ 
        \midrule
        \multirow{3}{*}{(Meta-)heuristics}
        & SJF 
        & 48.03 & 11.91\% & (1m) & 34.25 & 19.70\% & (1m) & 27.22 & 27.77\% & (1m)
        & 108.44 & 6.06\% & (2m) & 72.99 & 10.20\% & (2m) & 54.49 & 14.69\% & (3m) \\ 
        & GA 
        & 44.21 & 3.64\% & (2h) & 30.21 & 7.21\% & (2h) & 23.04 & 11.28\% & (2h) 
        & 104.51 & 2.37\% & (4h) & 68.91 & 2.47\% & (4h) & 50.48 & 7.08\% & (4h) \\ 
        & PSO
        & 43.84 & 2.80\% & (1.5h) & 29.79 & 5.82\% & (1.5h) & 22.70 & 9.79\% & (1.5h)
        & 104.70 & 2.55\% & (3h) & 69.13 & 4.79\% & (3h) & 50.71 & 7.53\% & (3h) \\ 
        \midrule  
        \multirow{4}{*}{Autoregressive}
        & MatNet  & 
        43.78 & 2.67\% & (0s) & 29.59 & 5.12\% & (0s) & 21.36 & 3.70\% & (1s) 
        & 103.16 & 1.07\% & (2s) & 67.09 & 1.79\% & (1s) & 47.67 & 1.36\% & (6s) \\
        & MatNet ($\times$8) & 
        42.87 & 0.56\% & (1s) & 28.49 & 1.35\% & (1s) & 20.87 & 1.41\% & (2s) 
        & 102.42 & 0.34\% & (5s) & 66.13 & 0.35\% & (3s) & 47.31 & 0.59\% & (17s) \\
        & MatNet ($\times$32) & 
        42.73 & 0.24\% & (2s) & 28.35 & 0.86\% & (2s) & 20.81 & 1.12\% & (6s) 
        & 102.23 & 0.16\% & (9s) & 66.11 & 0.31\% & (16s) & 47.19 & 0.33\% & (39s) \\
        & MatNet ($\times$128)& 
        42.67 & 0.10\% & (9s) & 28.28 & 0.62\% & (15s) & 20.68 & 0.50\% & (37s)
        & 102.14 & 0.08\% & (42s) & 66.02 & 0.18\% & (1m) & 47.12 & 0.18\% & (5m) \\
        \midrule 
        \multirow{2}{*}{Heatmap-based}
        & SL ($\times$128) & 
        45.31 & 6.09\% & (1m) & 31.12 & 10.16\% & (2m) & 23.76 & 14.34\% & (2m) 
        & 107.88 & 5.54\% & (3m) & 71.95 & 8.78\% & (3m) & 53.65 & 13.15\% & (4m) \\
        & RL ($\times$128) & 
        63.94 & 39.997\% & (9m) & 50.93 & 57.74\% & (9m) & 43.39 & 71.323\% & (10m) & 
        152.86 & 39.858\% & (22m) & 120.08 & 58.26\% & (24m) & 107.719 & 78.435\% & (28m) \\
        \midrule 
        \multirow{4}{*}{Ours}
        & IC/DC  & 
        43.06 & 1.00\% & (3s) & 29.17 & 3.69\% & (7s) & 21.16 & 2.80\% & (4s) & 
        102.87 & 0.79\% & (14s) & 66.68 & 1.17\% & (10s) & 48.18 & 2.41\% & (15s) \\
        & IC/DC ($\times$8) & 
        42.67 & 0.16\% & (5s) & 28.30 & 0.67\% & (15s) & 20.75 & 0.80\% & (7s) & 
        102.27 & 0.21\% & (20s) & 66.20 & 0.45\% & (28s) & 47.44 & 0.87\% & (45s) \\
        & IC/DC ($\times$32) & 
        \textbf{42.64} & \textbf{0.02\%} & (21s) & 
        \textbf{28.19} & \textbf{0.28\%} & (46s) & 
        \textbf{20.66} & \textbf{0.39\%} & (25s) & 
        \textbf{102.13} & \textbf{0.06\%} & (50s) & 
        66.06 & 0.24\% & (1m) & 
        47.22 & 0.39\% & (2m) \\
        & IC/DC ($\times$128)& 
        \textbf{42.63} & \textbf{0\% }& (47s) & 
        \textbf{28.15} & \textbf{0.14\%} & (3m) & 
        \textbf{20.61} & \textbf{0.15\%} & (2m) & 
        \textbf{102.08} & \textbf{0.01\%} & (3m) & 
        \textbf{65.97} & \textbf{0.11\%} & (5m) & 
        \textbf{47.11} & \textbf{0.17\%} & (7m) \\
        \bottomrule
    \end{tabular}
    \caption{\textbf{(Results for PMSP-20 and PMSP-50)} The table shows Makespan (M.S.), Gap, and Time for 3 × 20, 4 × 20, and 5 × 20, along with PMSP-50 (3 × 50, 4 × 50, 5 × 50).}
    \label{table:PSMP result}
\end{table*}

\begin{table*}[ht] 
    \centering
    \setlength{\tabcolsep}{5pt} 
    \begin{tabular}{llcccccc}
        \toprule
        & & \multicolumn{3}{c}{ATSP-20} & \multicolumn{3}{c}{ATSP-50} \\
        \cmidrule(lr){3-5}\cmidrule(lr){6-8}
        & & Tour Length $\downarrow$ & Gap $\downarrow$ & Time $\downarrow$ & Tour Length $\downarrow$ & Gap $\downarrow$ & Time $\downarrow$ \\
        \midrule
        Baseline & CPLEX & 1.534 & 0\% & (2m) & 1.551 & 0\% & (2h) \\ 
        \midrule
        \multirow{4}{*}{(Meta-)heuristics} & Nearest Neighbor & 1.994 & 26.099\% & (0s) & 2.092 & 29.701\% & (0s) \\
        & Nearest Insertion & 1.791 & 15.477\% & (0s) & 1.938 & 22.164\% & (0s) \\
        & Furthest Insertion & 1.709 & 10.770\% & (0s) & 1.836 & 16.802\% & (0s) \\
        & LKH-3 & 1.561 & 1.758\% & (1s) & 1.551 & 0\% & (8s) \\ 
        \midrule
        \multirow{4}{*}{Autoregressive}& MatNet & 1.541 & 0.456\% & (0s) & 1.572 & 1.314\% & (1s) \\
        & MatNet ($\times 8$) & 1.535 & 0.078\% & (2s) & 1.556 & 0.317\% & (15s) \\
        & MatNet ($\times 32$) & 1.534 & 0.033\% & (9s) & 1.554 & 0.214\% & (31s) \\
        & MatNet ($\times 128$) & 1.534 & 0.013\% & (37s) & \textbf{1.553} & \textbf{0.111}\% & (2m) \\ 
        \midrule
        \multirow{2}{*}{Heatmap-based}& Diffusion (SL) ($\times 128$) & $1.599$ & $4.179\%$ & (5m) & $1.684$ & $8.215\%$ & (30m) \\ 
        & Diffusion (RL) ($\times 128$) & 3.331 & $73.875\%$ & (6m) & 4.334 & $94.579\%$ & (32m) \\ 
        & Diffusion (SL+RL) ($\times 128$) & 1.589 & $3.508\%$ & (5m) & 1.679 & $7.964\%$ & (30m) \\ 
        \midrule
        \multirow{4}{*}{Ours}& IC/DC & 1.609 & 4.802\% & (1s) & 1.619 & 4.266\% & (15s) \\ 
        & IC/DC ($\times 8$) & 1.548 & 0.909\% & (8s) & 1.570 & 1.243\% & (2m) \\ 
        & IC/DC ($\times 32$) & 1.546 & 0.779\% & (31s) & 1.564 & 0.828\% & (8m) \\ 
        & \textbf{IC/DC} ($\times 128$) & \textbf{1.534} & \textbf{0}\% & (2m) & 1.553 & 0.113\% & (20m) \\ 
        \bottomrule
    \end{tabular}
    \caption{\textbf{(Results for ATSP-20 and ATSP-50)} The top row (labeled as baseline) represents the results from an off-the-shelf solver and is used as a reference for calculating the performance gap.
} 
    \label{table:ATSP result} 
\end{table*}


\subsection{Demonstrated Problems}
We begin by describing the combinatorial optimization (CO) problems on which we conducted experiments: the Parallel Machine Scheduling Problem (PMSP) and the Asymmetric Travelling Salesman Problem (ATSP).
\vspace{-8pt}
\paragraph{Parallel machine scheduling problem}

In PMSP, a problem instance \( c \) consists of \( |\mathcal{J}| \) jobs and \( |\mathcal{M}| \) machines. Each job \( j \in \mathcal{J} \) must be scheduled on a machine \( m \in \mathcal{M} \), with varying workloads for each job and different processing capabilities for each machine. 
The primary objective in PMSP is to minimize the makespan, which is the total length of the schedule upon the completion of all jobs. 
In this context, having $[X_0]_{j,m}=1$ indicates that job $j$ is assigned to machine $m$, which takes a processing time of $[P]_{j,m}$ where $P\in \mathbb{R}_+^{|\mathcal{J}| \times |\mathcal{M}|}$ is a matrix of processing times for all combinations.
The goal is to determine the solution matrix \( X_0 = \{ 0, 1\}^{|\mathcal{J}| \times |\mathcal{M}|} \) that minimizes the makespan for a given problem \( c=(\mathcal{M}, \mathcal{J}, P)\):
\begin{equation}
    \text{score}_{\text{PMSP}}(X_0,c) = -\max_m \sum_j [X_0\odot P]_{j,m}.
\end{equation}
The solution matrix that assigns a job to multiple machines is considered infeasible.
\vspace{-8pt}
\paragraph{Asymmetric travelling salesman problem}

An ATSP instance $c=(|\mathcal{N}|, D)$ comprises $|\mathcal{N}|$ cities and an asymmetric distance matrix $D\in\mathbb{R}_+^{|\mathcal{N}|\times |\mathcal{N}|}$ where each element of it specifies the distance between two cities.
The solution to ATSP is a tour, which is an adjacency matrix $X_0 = \{ 0, 1\}^{|\mathcal{N}| \times |\mathcal{N}|}$ for a directed graph visiting all cities once. The goal is find a solution that minimizes the tour length:
\begin{equation}
    \text{score}_{\text{ATSP}}(X_0,c) = -\sum_{i, j}{[X_0 \odot D]_{i, j}}.
\end{equation}
The solution matrix that is not a Hamiltonian cycle is considered infeasible.
In accordance with~\citet{kwon2021matrix} we employ tmat-class ATSP instances (see~\cref{Tmat class}).

\begin{figure*}[t]
\centering
\includegraphics[width=1.0\linewidth]{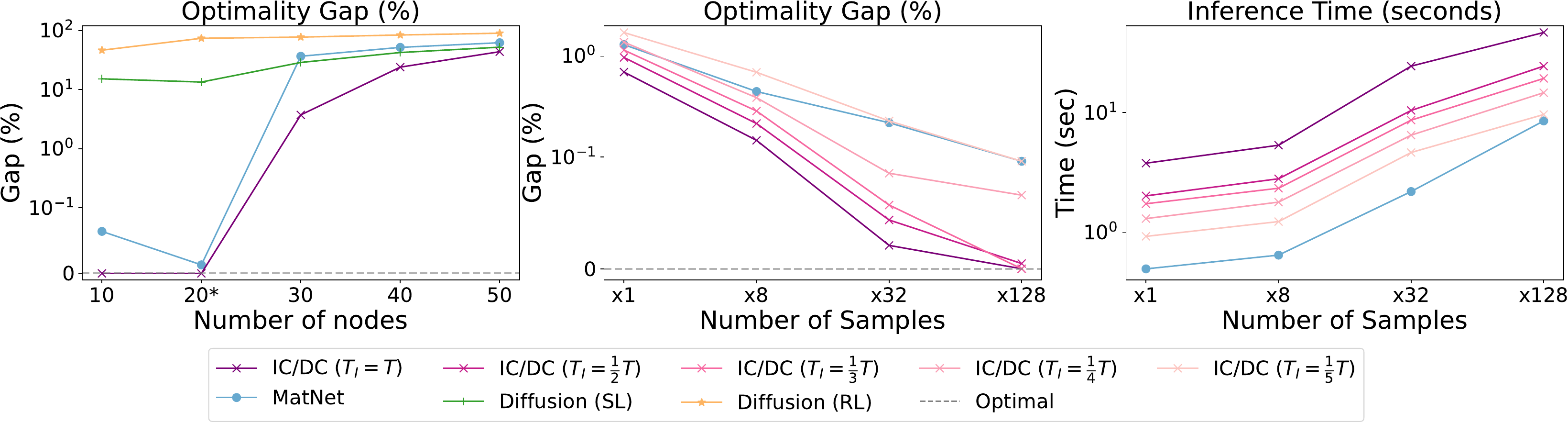} 
\caption{(\textbf{Left}) shows the optimality gap among different methods in ATSP. Each of the methods used a unified sampling size of 128. The * indicates the training distribution (20 nodes). (\textbf{Center}) shows the optimality gap across sampling step size in PMSP (3 machines and 20 jobs). (\textbf{Right}) shows the inference time across sampling step size.}
\label{fig:generalize_reduce_sample}
\end{figure*}

\subsection{Main Results}
For the evaluation, 1000 problem instances were randomly generated using a standard generation process (see~\cref{Tmat class}). To fully leverage the stochastic nature of generative learning-based methods, we also evaluated these methods by generating multiple samples ($\times n$) for each problem instance and selecting the one with the best score. For MatNet~\citep{kwon2021matrix}, we followed the authors' implementation, including instance augmentation, which yielded better result. 
As a problem-specific search process has not been studied on PMSP and ATSP, for heatmap-based methods, we report simple discrete diffusion models, either trained with supervised learning~\citep{sun2023difusco} or with reinforcement learning~\citep{black2023training}. For further experimental details, please refer to~\cref{PMSP_detail} and~\cref{ATSP_detail}.

\paragraph{PMSP}
We evaluated our method on PMSP-20 and PMSP-50, where the numbers 20 and 50 correspond to the number of jobs in each instance, with the number of machines fixed at 4. Our approach is compared against various baselines, including (meta-)heuristics, autoregressive, and heatmap-based methods. Details of these baselines can be found in~\cref{pmsp baseline}.
As shown in~\cref{table:PSMP result}, IC/DC achieves the smallest optimality gap among learning-based methods. IC/DC demonstrates a $0.142\%$ performance gap compared to CP-SAT, whereas the previous SOTA, MatNet, shows a $0.615\%$ gap on PMSP-20. On PSMP-50 IC/DC reduces the gap from MatNet's $0.182\%$ to $0.112\%$.

With combination of feasibility-enforced generation and heatmap-based iterative refinement, ID/DC achieves greater sample diversity, as the denoising process provides varied foundations for solutions before the autoregressive generation.
While IC/DC's performance lags behind MatNet when using a single generation to solve the problem, it quickly surpasses MatNet as the number of samples increases~(\cref{table:PSMP result}). Upon closer examination, we observed that the stochasticity of MatNet primarily stems from the starting point of the autoregressive generation (i.e., the assignment of the initial job for scheduling). In contrast, IC/DC's denoising process offers diverse backbones for the autoregressive generation, resulting in varied samples even when starting from the same initial  assignment.
Although IC/DC has a slower inference speed compared to autoregressive methods, its high-quality solutions remain highly competitive with other baselines.
\vspace{-10pt}
\paragraph{ATSP}
We evaluated our method on ATSP-20 and ATSP-50, where the number 20 and 50 correspond to the number of cities. 
To our best knowledge, IC/DC is the first framework among NCO to achieve optimality gap with $0\%$ in ATSP-20.
This improvement in performance positions IC/DC as a new competitive learning-based method for addressing these less-studied CO problems, highlighting its potential.

As shown in~\cref{table:PSMP result} and~\cref{table:ATSP result}, simple implementations of diffusion models, whether trained with supervised learning~\citep{sun2023difusco} or reinforcement learning~\citep{black2023training}, struggle to generate high quality solutions without the aid of problem-specific search processes, which require significant expertise in the specific CO problem being addressed. 
This highlights the critical importance of our feasibility-enforced generation process, which integrates the flexibility of autoregressive methods into heatmap-based methods, allowing for the straightforward imposition of various constraints. In this regard, the proposed IC/DC approach significantly broadens the applicability of heatmap-based methods to a wide range of less-explored CO problems with diverse constraints.
\vspace{-2pt}
\subsection{Generalization capability}
To assess the ability to generalize of IC/DC, we train our model on the ATSP-20 setting and evaluate it on ATSP instances with different numbers of nodes. Left side of~\cref{fig:generalize_reduce_sample} shows the result of this evaluation, comparing IC/DC with other learning-based methods. IC/DC demonstrates superior generalization capabilities, achieving $0\%$ optimality gap for 10 nodes. As the number of nodes increases, IC/DC maintains a lower optimality gap, with a notable $3.733\%$ gap at 30 nodes and relatively low gaps at higher node counts compared to other methods. These results suggests that IC/DC can effectively adapt to a range of different problems, confirming its ability to generalize across diverse, unseen distributions. (See~\cref{generalizability_full_result} for full results)


\subsection{Reducing sampling steps for faster inference}
Compared to autoregressive methods, diffusion models exhibit slower inference time due to the expensive reverse process.
The parameterization in \cref{reverse_process_eq} enables inference with reduced sampling steps with a cost of accuracy through $p_\theta(X_{t-s}|X_t, c) = \sum_{X_0} q(X_{t-s}, X_t|X_0)p_\theta(X_0|X_t, c)$~\citep{austin2021structured}, setting $t>1$.
We examine the trade-offs associated with reduced sampling steps in the PMSP setting with 3 machines and 20 jobs. As shown in center and right of~\cref{fig:generalize_reduce_sample}, the inference time decreases significantly as the number of sampling steps during inference $T_I$ decreases. 
IC/DC attains a $0\%$ optimality gap for both $T_I=\frac{1}{2}T$ and $T_I=\frac{1}{3}T$ settings with competitive inference speed.
The trade-off reported here can be improved using a more sophisticated distillation algorithms~\citep{song2023consistency}. (See~\cref{Reducing_sampling_faster_inference_additional_results} for more results)

\vspace{-3pt}
\section{Conclusion}
\label{sec:conclusion}

We propose a diffusion model framework, IC/DC, to tackle the CO problems with two distinct sets of items, showing competitive performance in both PMSP and ATSP. Our algorithm outperforms existing NCO baselines and generates feasible solutions that are difficult for heatmap-based methods. We believe the proposed IC/DC framework is flexible, and can be broadly applied to various domains;
e.g., in~\cref{app_real}, we show its capability to address real-world CO problems with complex item and relationship features.

Despite its strong performance, IC/DC has a notable limitation: the GAT-based encoder we use requires $O(\max(|\mathcal{A}|, |\mathcal{B}|)^2)$ memory complexity, demanding significant resources for large instances. Unfortunately, due to limited computational resources, we were not able to assess the performance of IC/DC on larger problem instances, which prevented us from fully exploring IC/DC’s scalability potential. To address these challenges, we plan to explore memory-efficient techniques in future work, such as those introduced by~\citet{zhu2024scalable}.


{
    \small
    \bibliographystyle{ieeenat_fullname}
    \bibliography{main}
}

\clearpage
\setcounter{page}{1}
\newpage
\onecolumn
{\centering
\Large
\textbf{\thetitle}\\
\vspace{0.5em}Supplementary Material \\
\vspace{1.0em}}


\appendix

\section{Derivation}
\label{Derivation}

\subsection{Joint Variational Upper Bound with EUBO}\label{Joint Variational Upper Bound}
We derive the upper bound using the Evidence Upper Bound (EUBO)~\citep{ji2019stochastic}
\begin{align}
    \log \tilde{q}(X|c) &= \int \tilde{q}(Z|X,c) \log \tilde{q}(X|c) dZ\\ &= \int \tilde{q}(Z|X,c) [\log \tilde{q}(X, Z|c) - \log \tilde{q}(Z|X,c)] dZ\\ &\leq \int \tilde{q}(Z|X,c) [\log \tilde{q}(X, Z|c) - \log p_{\theta}(Z|X,c)] dZ
\end{align}

The inequality is established by the Gibbs' inequality: $-\int \tilde{q}(Z|X,c)\log \tilde{q}(Z|X,c) \leq -\int \tilde{q}(Z|X,c)\log p_\theta (Z|X,c)$.
\begin{align}
    D_{KL}(\tilde{q}(X|c) \| p_{\theta}(X|c)) &= \int \tilde{q}(X|c) \log \frac{\tilde{q}(X|c)}{p_{\theta}(X|c)} dX \\
    &\leq \int \tilde{q}(X|c) \left[ \int \tilde{q}(Z|X,c) \left( \log \tilde{q}(Z,X|c) - \log p_{\theta}(Z|X,c) \right) dZ - \log p_{\theta}(X|c) \right] dX \\
    &\leq \int \int \tilde{q}(X, Z|c) [\log \tilde{q}(X, Z|c)-\log p_{\theta}(Z|X,c) - \log p_{\theta}(X|c)] dZ dX \\
    &= D_{KL}(\tilde{q}(X,Z|c)||p_{\theta}(X,Z|c))
\end{align}
The last inequality is derived from $\int \tilde{q}(X|c) \log p_{\theta}(X|c) dX = \int \tilde{q}(X,Z|c) \log p_{\theta}(X,Z|c) dZ dX$. 

\subsection{KL divergence to Reward-weighted Diffusion Loss}\label{reward weighed diffusion loss}
We derive the reward-weighted diffusion loss from the objective in \texttt{CLONING} step. We aim to minimize the Kullback-Leibler (KL) divergence $D_{KL}(\tilde{q}(X|c) \parallel p_\theta(X|c))$ between the surrogate target and the distribution \(p\) parameterized by \(\theta\). however since evaluating the log-likelihood of the generative model \(p_{\theta}(X|c)\) is difficult~\citep{kingma2013auto}, we instead utilize a joint variational upper bound (see~\cref{Joint Variational Upper Bound}):
\begin{equation}\label{Inequality}
       \mathrm{D}_{\mathrm{KL}}(\tilde{q}(X|c) \parallel p_\theta(X|c)) \leq \mathrm{D}_{\mathrm{KL}}(\tilde{q}(X, Z|c) \parallel p_\theta(X, Z|c))
\end{equation}
According to Gibbs' inequality if the right-hand side of the inequality is zero, the inequality becomes an equality. In this case, the surrogate target is exactly approximated.

Therefore, we focus on minimizing the right-hand side of eq~\eqref{Inequality}. By applying the diffusion process where \( Z = X_{1:T} \), \( X = X_0 \), and \( T \) is the diffusion step, the forward process is defined by $q(X_{1:T}|X_0)$ with $X_0 \sim \tilde{q}(X_0|c)$.
We minimize the Reward-weighted Diffusion Loss $\mathcal{L}_{\text{RWD}}$:
\begin{align}
    \mathcal{L}_{\text{RWD}} &= D_{KL}(\tilde{q}(X_{0:T}|c) \parallel p_\theta(X_{0:T}|c)) \\
    &= D_{KL}(\tilde{q}(X_0,X_{1:T}|c) \parallel p_\theta(X_{0:T}|c))\\
    &= \mathbb{E}_{X_0 \sim \tilde{q}(X_0|c), X_{1:T} \sim q(X_{1:T}|X_0)}\left[ \log \frac{q(X_{1:T}|X_0)\tilde{q}(X_0|c)}{p_{\theta}(X_{0:T}|c)}\right] \\
    &= \mathbb{E}_{X_0 \sim \tilde{q}, X_{1:T} \sim q}\left[\log \frac{q(X_{1:T}|X_0)}{p_{\theta}(X_{0:T}|c)} + \log \tilde{q}(X_0|c) \right] \\
    &= \mathbb{E}_{X_0 \sim \tilde{q}, X_{1:T} \sim q} 
    \left[\log \frac{\prod_{t=1}^{T} q(X_t|X_{t-1})}{p_\theta(X_T|c) \prod_{t=1}^{T} p_\theta(X_{t-1}|X_t, c)} + \log \tilde{q}(X_0|c) \right] \\
    &= \mathbb{E}_{X_0 \sim \tilde{q}, X_{1:T} \sim q}
    \left[ -\log p_\theta(X_T|c) + \sum_{t=1}^{T} \log \left( \frac{q(X_t|X_{t-1})}{p_\theta(X_{t-1}|X_t, c)} \right) +\log \tilde{q}(X_0|c) \right] \\
    &= \mathbb{E}_{X_0 \sim \tilde{q}, X_{1:T} \sim q} 
    \left[ -\log p_\theta(X_T|c) + \sum_{t=2}^{T} \log \left( \frac{q(X_t|X_{t-1})}{p_\theta(X_{t-1}| X_t, c)} \right) + \log \left( \frac{q(X_1|X_0)}{p_\theta(X_0|X_1, c)} \right) + \log \tilde{q}(X_0|c)\right] \\
    &= \mathbb{E}_{X_0 \sim \tilde{q}, X_{1:T} \sim q} 
    \Bigg[-\log p_\theta(X_T| c) \\
    &\quad\quad\quad\quad\quad\quad\quad \ \ + \sum_{t=2}^{T} \log \left( \frac{q(X_{t-1}|X_{t}, X_0)}{p_\theta(X_{t-1}|X_t, c)} \cdot \frac{q(X_t|X_0)}{q(X_{t-1}|X_0)} \right) + \log \left( \frac{q(X_1|X_0)}{p_\theta(X_0|X_1, c)} \right) +\log \tilde{q}(X_0|c)\Bigg] \\
    &= \mathbb{E}_{X_0 \sim \tilde{q}, X_{1:T} \sim q} 
    \left[\log \frac{q(X_T|X_0)}{p_\theta(X_T|c)} + \sum_{t=2}^{T} \log \left( \frac{q(X_{t-1}|X_{t}, X_0)}{p_\theta(X_{t-1}|X_t, c)} \right) - \log p_\theta(X_0|X_1, c) + \log \tilde{q} (X_0|c) \right] \\
    &= \mathbb{E}_{X_0 \sim \tilde{q}, X_{1:T} \sim q} \Bigg[\underbrace{D_{KL}(q(X_T|X_0) \| p_\theta(X_T|c))}_{L_T} + \sum_{t=2}^{T} \underbrace{D_{KL}(q(X_{t-1}|X_{t}, X_0) \| p_\theta(X_{t-1}|X_t, c))}_{L_{t-1}} \\
    &\quad\quad\quad\quad\quad\quad\quad \ \ + \underbrace{D_{KL}(\tilde{q}(X_0|c) \| p_\theta(X_0|X_1, c))}_{L_0}\Bigg] 
\end{align}

If $T$ is sufficiently large, $L_T$ will approach zero~\citep{austin2021structured}. Also, $L_0$ can be derived as CE loss:
\begin{align}
    L_0 &= \sum \tilde{q}(X_0|c) \log \frac{\tilde{q}(X_0|c)}{p_\theta(X_0|X_1, c)} = \sum \tilde{q}(X_0|c) \log \tilde{q}(X_0|c) - \log p_\theta(X_0|X_1, c)) \\
    &= CE(\tilde{q},p_\theta) - H(\tilde{q}) \approx \mathcal{L}_{\text{prd}} -H(\tilde{q})
\end{align}
Cross-entropy loss is equal to $\mathcal{L}_{\text{prd}}$ at $t=1$. As $H(\tilde{p})$ is independent to $p_\theta$, it can be ignored for training. As $f_\theta$ predicts the logit of $X_0$, we use the Cross-entropy loss for every timestep $L_\text{pred}$ instead of Cross-entropy loss at $t=1$.

\begin{align}
    \therefore \mathcal{L}_{\text{RWD}} \approx \mathcal{L}_{\text{VB}}+\mathcal{L}_{\text{prd}}
\end{align}
Training with this loss function results in an enhancement of solution quality.

\subsection{Constraint Loss}\label{constraint loss}
In CO problems, solutions must strictly satisfy specific constraints. To discourages infeasible predictions $X_0 \sim p_\theta(X_0|X_t, c)$, we introduce a constraint loss:
\begin{align}
\mathcal{L}_{\text{cst}}(\theta)&=\mathbb{E}_{X_0\sim \tilde{q}(X_0|c), X_t\sim q(X_t|X_0)}\left[C( p_\theta(X_0|X_t, c))\right],
\end{align}

where $C(\cdot)$ is a differentiable function that approximately measures constraint violations of samples

In the Asymmetric Travelling Salesman Problem (ATSP), each node is required to travel to a distinct city other than itself. Consequently, in the resulting solution matrix, each row and each column must contain exactly one entry of '1':

\begin{equation}
    C_{\text{ATSP}}(X_0) = \left( \sum_{i} \text{Gumbel-Softmax}(X_0)_{i,j,1} - 1 \right)^2 + \left( \sum_{j} \text{Gumbel-Softmax}(X_0)_{i,j,1} - 1 \right)^2
\end{equation}
In the Parallel Machine Scheduling Problem (PMSP), each job is required to be assigned to a single machine. Accordingly, in the solution matrix, each column must contain exactly one entry of '1', with all other entries in that column being '0':

\begin{equation}
    C_{\text{PMSP}}(X_0) = \left( \sum_{j} \text{Gumbel-Softmax}(X_0)_{i,j,1} - 1 \right)^2
\end{equation}
where $\text{Gumbel-Softmax}(x)_i = \frac{\exp((x_i + g_i) / \tau)}{\sum_{j=1}^2 \exp((x_j + g_j) / \tau)}$, and \(g_i\) is a value sampled from the Gumbel(0,1) distribution, and \(\tau\) is the temperature parameter.

\subsection{Diffusion Loss}\label{diffusion objective}
By combining reward-weighted diffusion loss~\cref{reward weighed diffusion loss} and constraint loss~\cref{constraint loss}, our diffusion objective:
\begin{equation}
    \mathcal{L}_{\texttt{CLN}}(\theta) = \mathbb{E}_{\tilde{q}} 
\left[\left(\sum_{t=2}^{T} D_{KL}(q(X_{t-1}|X_{t}, X_0) \| p_\theta(X_{t-1}|X_t, c)) -\lambda_1 p_\theta(X_0|X_t, c)\right)+ \lambda_2 C(p_\theta(X_{0}|X_t,c)) \right] 
\end{equation}
where $\lambda_1$ and $\lambda_2$ are hyper parameter.

\section{IC/DC in detail}
\subsection{Choice of Transition Matrices}\label{qbar_formula}
According to~\citep{austin2021structured}, it is argued that incorporating domain-specific structures into the transition matrices \( Q_t \) within the diffusion process is a reasonable approach. In our case, due to the inherent sparsity in the solution matrices of CO problems, the marginal distribution of feasible solutions significantly deviates from the uniform distribution commonly used in standard diffusion processes. Therefore, we design this noise transition matrix to align with the prior distribution of feasible solutions \( \bar{q} \).

Depending on what CO problem we are aiming to solve, we are often able to compute the prior distribution, averaged over solution elements $\bar{q}(\tilde{x})=\bar{q}\left(\frac{1}{|\mathcal{A}||\mathcal{B}|}\sum_{i=1}^{|\mathcal{A}||\mathcal{B}|}[\tilde{x}_0]_i\right)$.
The marginal probability of $x=1$ over a set of feasible solutions can be expressed as $\bar{q}(x=1)$. 

In ATSP, when considering \(|\mathcal{N}|\) and $N_{\text{col}}$ cells, the solution involves travelling through all \(|\mathcal{N}|\) cities. The prior distribution follows the following:
\begin{equation}
    \bar{q}= [1-\frac{1}{|\mathcal{N}|}, \frac{1}{|\mathcal{N}|}]^\top
\end{equation}

In PMSP, when considering \(|\mathcal{J}|\) jobs and \(|\mathcal{M}|\) machines, the solution involves assigning all \(|\mathcal{J}|\) jobs to the machines. The prior distribution is as follows:
\begin{equation}
    \bar{q}= [1-\frac{1}{|\mathcal{J}|}, \frac{1}{|\mathcal{J}|}]^\top
\end{equation}
In most combinatorial optimization problems, the marginal distribution $\bar{q}$ is typically known. However, in cases where the marginal distribution is not available, an alternative approach is to utilize a uniform distribution. As a substitute for $\bar{q}$, one may consider using $\bar{u} = [0.5, 0.5]$.

\subsection{Problem Encoder}
\label{sec:problem_encoder}
Denoting each row of $A$ as $A=[a_1,...,a_{|\mathcal{A}|}]^\top$, an attention block within each layer processes the input as:
\begin{align}
S_{\text{inter}} &= \text{softmax}(A W_{\text{inter}}A^{\top})&\in \mathbb{R}^{|\mathcal{A}| \times |\mathcal{A}|}, \\
S_{\text{intra}} &= \text{ReLU}(A W_{\text{intra}}{B}^{\top}) &\in \mathbb{R}^{|\mathcal{A}| \times |\mathcal{B}|}, \\
S&=\text{concat}(S_{\text{inter}}S_{\text{intra}}, D)&\in \mathbb{R}^{|\mathcal{A}| \times |\mathcal{B}|\times 2},\\
\tilde{D} &= \text{MLP}(S) &\in \mathbb{R}^{|\mathcal{A}| \times |\mathcal{B}|}, \\
\tilde{A} &= \text{softmax}(\tilde{D}) BW_{\text{v}}&\in \mathbb{R}^{|\mathcal{A}| \times d},
\end{align}
where $W_{\text{inter}}$, $W_{\text{intra}}$, and $W_{\text{v}}$ are weight matrices of dimension $\mathbb{R}^{d\times d}$, and $\text{MLP}(\cdot)$ is a fully-connected neural network that maps $2$-dimensional inputs to $1$-dimensional outputs. The matrices $S_{\text{inter}}$ and $S_{\text{intra}}$ are designed to capture the inter-relationships within set $\mathcal{A}$ and the intra-relationships and between sets $\mathcal{A}$ and $\mathcal{B}$. These matrices, along with $D$, are combined to form the attention score $\tilde{D}$, which produces the output $\tilde{A}$. Using the described attention block, the layer outputs $A'$ as follows:
\begin{equation}
    \hat{A}=\text{BN}(A+\tilde{A}),\quad A'=\text{BN}(\hat{A} + \text{MLP}(\hat{A})),
\end{equation}
where $\text{BN}$ refers to a batch normalization~\citep{ioffe2015batch}. The process for updating $B$ to $B'$ is computed in the same way.

\subsection{Denoiser}
\label{sec:denoiser}
As illustrated in the bottom-left side of ~\cref{fig:architecture}, the denoiser consists of $L'$ layers, where each layer gets input of $(A, B, X, t)$ and outputs $(A', B', X')$, which is processed as follows:

\begin{align}
\hat{x}_{i, j}^{\ell+1} &= P^{\ell} x_{i, j}^{\ell} + Q^{\ell} h_{i}^{\ell} + R^{\ell} h_{j}^{\ell}, \\    
x_{i, j}^{\ell+1} &= x_{i, j}^{\ell} + \text{MLP}_x(\text{BN}(\hat{x}_{i, j}^{\ell+1})) + \text{MLP}_t(\bold{t}), \\    h_{i}^{\ell+1} &= h_i^{\ell} + \text{ReLU} \left( \text{BN}(U^{\ell}_a h_{i}^{\ell} + \sum_{j \in \mathcal{N}_i} (\sigma(\hat{x}_{i, j}^{\ell+1}) \odot V^{\ell}_{b} h_{j}^{\ell})) \right), \\
h_{j}^{\ell+1} &= h_j^{\ell} + \text{ReLU} \left( \text{BN}(U^{\ell}_{b} h_{j}^{\ell} + \sum_{i \in \mathcal{N}_j} (\sigma(\hat{x}_{j, i}^{\ell+1}) \odot V^{\ell}_a h_{i}^{\ell})) \right),
\end{align}
Where  \( h^{\ell=0}_i = h^L_i \) and \( h^{\ell=0}_j = h^L_j \). For simplicity, the vector representation \( x^{\ell=0}_{i,j} \) at the \((i,j)\)-th \( X_t \) is denoted without \( t \).

The matrices \( U^\ell_{a}, U^\ell_{b}, V^\ell_{a}, V^\ell_{b}, P^\ell, Q^\ell, R^\ell \in \mathbb{R}^{d \times d} \) are learnable parameters of the \(\ell\)-th layer. SUM pooling is denoted by \(\sum\)~\citep{xu2018powerful}, the sigmoid function is represented by \(\sigma\), and the Hadamard product is denoted by \(\odot\). \( \mathcal{N}_i \) denotes the neighborhood of node \( i \) among the $B$ items, while \( \mathcal{N}_j \) denotes the neighborhood of node \( j \) among the $A$ items. Additionally, the variable \(\bold{t}\) denotes the sinusoidal features~\citep{vaswani2017attention} corresponding to the denoising timestep $t$. 
After the final layer \( \mathfrak{L} \), the output \( x^{\mathfrak{L}}_{i,j} \) is passed through a linear layer to obtain clean matrix \( X_0 = \{x_{0,0}, ..., x_{I, J} \} \).

\section{PMSP definition and baselines}\label{PMSP_detail}
\subsection{Unrelated Parallel Machine}
In the literature on the parallel machine scheduling problem, most studies focus on a simplified variant where the processing time for a job is consistent across all machines, commonly referred to as "uniform parallel machines". In contrast, we examine a more complex scenario where the processing times for each machine are entirely independent to one another. For our study, we generate the processing time matrix randomly and use it as our instance.

\subsection{Baseline}\label{pmsp baseline}
\paragraph{Mixed-integer programming}
Constraint Programming with Satisfiability (CP-SAT) is a highly efficient solver developed as part of the OR-Tools suite, designed for solving integer programming problems. It is particularly effective for solving scheduling problems, including PMSP.
For PMSP our MIP model is based on \citet{avalos2013reformulation}
\begin{align}
    &\text{minimize} \quad C_{max} \\
    \text{s.t.}\quad
    &C_0 = 0 \label{pmsp_const6} \\
    &\sum_{j \in J} x_{i0j} \leq 1 \quad\quad i \in M \label{pmsp_const5} \\
    &\sum_{j \in J_0, j \neq k}\sum_{i \in M}x_{ijk} = 1 \quad\quad k \in J \label{pmsp_const1} \\
    &\sum_{k \in J_0, j \neq k}\sum_{i \in M}x_{ijk} = 1 \quad\quad j \in J \label{pmsp_const2} \\
    &\sum_{j \in J_0, j \neq k} \sum_{k \in J}p_{ik}x_{ijk} \leq C_{max} \quad i \in M \label{pmsp_const7} \\
    &\sum_{k \in J_0, j \neq k}x_{ijk} = \sum_{h \in J_0, h \neq j}x_{ihj} \quad\quad j \in J, i \in M  \label{pmsp_const3} \\
    &C_k - C_j + V(1 - x_{ijk}) \geq p_{ik} \quad j \in J_0, k \in J, j \neq k, i \in M \label{pmsp_const4} 
\end{align}
\begin{align}
    &\text{where} \\
    &C_{max} : \text{Maximum completion time (makespan) (MS)} \\
    &C_j : \text{Completion time of job j} \\
    &x_{ijk} : \begin{cases}
        1 & \text{if job $k$ is processed directly after job $j$ on machine $i$}\\
        0 & \text{otherwise}
    \end{cases}.\\
    &p_{ik}: \text{Time required to process job $j$ on machine $i$} \\
    &M: \text{Set of machines} \\
    &J: \text{Set of jobs to schedule} \\
    &J_0: \text{Set of jobs to schedule with an additional dummy node (indexed by 0)} \\
    &V: \text{A large positive number}
\end{align}
Constraints~\eqref{pmsp_const1} and \eqref{pmsp_const2} ensure that each job has exactly one predecessor on one of the machines. Constraint\eqref{pmsp_const3} specifies that if a job has a predecessor on a machine, it must also have a successor on that same machine. Constraint~\eqref{pmsp_const4} guarantees that a valid sequence of jobs is scheduled on each machine, with no overlap in processing times. Constraint~\eqref{pmsp_const5} ensures that only one job can be scheduled as the first job on each machine. Constraint~\eqref{pmsp_const6} sets the completion time of job 0, an auxiliary job used to define the start of the schedule, to zero. Finally, constraint~\eqref{pmsp_const7} establishes the relationship between the makespan of individual machines and the overall schedule makespan. CP-SAT is run on CPUs.

\paragraph{(Meta-)Heuristics}
Random and Shortest Job First (SJF) are greedy-selection algorithms designed to generate valid schedules using the Gantt chart completion strategy. SJF specifically prioritizes tasks by scheduling the shortest available jobs in ascending order at each time step $t$. Although simple, these methods can serve as effective baselines, providing a comparison point for evaluating more sophisticated scheduling algorithms. The greedy-selection algorithms run on CPUs.

CO problems often require sophisticated methods to find high-quality solutions, particularly when traditional approaches like Mixed Integer Programming (MIP) are impractical due to complexity. In such cases, meta-heuristics provide a robust alternative.  Genetic Algorithm (GA) and Particle Swarm Optimization (PSO) are two widely adopted meta-heuristics, known for their versatility and effectiveness across a range of problem domains.

GA iteratively updates multiple candidate solutions, referred to as chromosomes. New child chromosomes are produced by combining two parent chromosomes through crossover methods, and mutations are applied to the chromosomes to enhance exploration.

PSO iteratively updates multiple candidate solutions, referred to as particles. For every iteration each particles are updated based on the local best known and the global best known particles. 

We utilize the implementations provided by Kwon et al.~\citep{kwon2021matrix} and follow their setting. Both GA and PSO run on GPUs.

\paragraph{MatNet}
MatNet, proposed by~\citep{kwon2021matrix}, adapts the attention model~\citep{kool2018attention} 
to be applicable to bipartite graphs. It employs cross-attention mechanisms in place of self-attention to facilitate message passing between the two set of nodes, thereby encoding relationship information more effectively. The model is trained using reinforcement learning. Solution are generated in an auto-regressive manner, where each node is sequentially selected based on the current state of the solution. 

\paragraph{Diffusion (SL)}

We trained the diffusion model to estimate solution matrices using a supervised learning approach, similar to Difusco~\citep{sun2023difusco}. This method adapts graph-based denoising diffusion models to more naturally formulate combinatorial optimization problems and generate high-quality solutions. By explicitly modeling the node and edge selection process through corresponding random variables, the model effectively captures the problem's structure. The model is trained through supervised learning.


\paragraph{Diffusion (RL)}
Similar to DDPO, as proposed by ~\citep{black2023training}, which demonstrates that framing the denoising process as a multi-step decision-making problem enables policy gradient algorithms to directly optimize diffusion models for downstream objectives, we trained our diffusion model within the reinforcement learning framework. We used the REINFORCE algorithm~\citep{williams1992simple} to map the denoising process to the MDP framework.

\subsection{Training Detail}\label{pmsp train detail}

(Experimental detail)
We evaluated the baselines and the proposed algorithm using two Intel Xeon Gold 6330 CPUs and an RTX 3090 GPU for both PMSP and ATSP.

\paragraph{Genetic algorithm (GA)}
Following the implementation of~\citep{kwon2021matrix}, we utilize 25 chromosomes with a mutation rate and crossover ratio both set to 0.3. Among the 25 initial chromosomes, one is initialized with the solution from the SJF heuristic. The best-performing chromosome is retained across all iterations. We run 1000 iterations per instance.
\paragraph{Particle swam optimization (PSO)}
Following the implementation of ~\citep{kwon2021matrix}, we utilize 25 particles with an inertial weight of 0.7. Both the cognitive and social constants are set to 1.5. Additionally, one particle is initialized with the solution from the SJF heuristic. We run 1000 iterations per instance.
\paragraph{MatNet}
We use the same hyperparameters reported in ~\citep{kwon2021matrix}, with the exception that the number of stages is set to 1 instead of 3.

\paragraph{Diffusion (SL)}
Using CP-SAT, we generate 128,000 training samples for supervised learning.  Since the solution matrices for PMSP are not square matrices, we use IC/DC's encoder, employing 5 layers. The denoising timestep is set to 20. To generate a feasible solution, we apply greedy decoding after the denoising process. The greedy decoding selects the maximum probability for each job based on a heatmap generated by the model. If there is more than one machine with the maximum probability, one of them is chosen randomly. All other hyperparameters, follow those reported in ~\citep{sun2023difusco}

\paragraph{Diffusion (RL)}
We train standard discrete diffusion model with the objective function \\ $\nabla_\theta \mathcal{J}_{\text{DDRL}}=\mathbb{E}\left[ \sum_{t=0}^{T}\nabla_\theta\log p_\theta(X_{t-1}|X_t,c)r(X_0,c) \right]$ proposed by~\citep{black2023training}.We use the same encoder architecture as IC/DC, along with a GAT-based decoder, consisting of 5 encoder layers and 2 decoder layers.Additionally, we set the denoising timestep to 20. To enforce the model to output feasible solution, we apply greedy decoding, which is the same as the one used in Diffusion (SL). All other hyperparameters are consistent with those used in IC/DC.


\paragraph{IC/DC}
For PMSP-20 instances, we use 3 layers for both the encoder and the decoder, with the denoising timestep set to 10. For PMSP-50 instances, we utilize 5 encoder layers and 3 decoder layers, with the denoising timestep set to 15. The hyperparameters $lambda_1$ and $\lambda_2$ are set to 1e-3 and 1e-6, respectively. The \texttt{IMPROVEMENT} step is executed every 30 epochs. Our model is trained using the Adam optimizer, with a batch size of 512 and a learning rate of 4e-4.\\
Our code is available at: 
\begin{verbatim}
https://anonymous.4open.science/r/ICDC_opti-8347
\end{verbatim}

\section{ATSP definition and baselines}\label{ATSP_detail}
\subsection{Tmat class}\label{Tmat class}
While random distance matrices can be generated by choosing random integers, such matrices lack meaningful correlations between distances and doesn't reflect practical scenarios. Instead we are interested in problems which ATSP instances have the triangle inequality so called "Tmat class"~\citep{kwon2021matrix, cirasella2001asymmetric}. That is, for a distance matrix $D$ with elements $d_{ij}$ representing the distance between city $c_i$ and $c_j$, if $d(c_i, c_j) \geq d(c_i, c_k) + d(c_k, c_j)$ then we set $d(c_i, c_j) = d(c_i, c_j) + d(c_k, c_j)$, while diagonal elements are maintained as $d(c_i, c_i) = 0$. We repeat this procedure until no more changes can be made.
\subsection{Baseline}\label{atsp baseline}
\paragraph{Mixed-integer programming}
Mixed integer programming (MIP) is an optimization technique used to solve problems where some of the variables are required to be integers while others can be continuous. Solution methods such as branch and bound, branch and cut etc. are used to solve these kind of problems. To we use CPLEX~\citep{cplex, bliek1u2014solving}, one of the popular commercial optimization software use by the OR community and solve our test instances through benders decomposition~\citep{rahmaniani2017benders}.
The MIP model serves as the mathematical representation of the problem. For ATSP, our MIP model is based on the formulation presented by \citet{miller1960integer}.
\begin{align}
    &\text{minimize} \quad \sum^N_{i=1}\sum^N_{j=1}d(c_{ij}x_{ij}) \label{atsp_obj} \\
    \text{s.t.}\quad &\sum^N_{i=1}x_{ij} = 1 \quad\quad j = 1, 2, 3, \cdots, N \label{atsp_constraint1} \\
    &\sum^N_{j=1}x_{ij} = 1 \quad\quad i = 1, 2, 3, \cdots, N \label{atsp_constraint2} \\
    &u_i - u_j + (n-1) \cdot x_{ij} \leq n - 2 \quad i,j = 2, \cdots, N \label{atsp_constraint3}
\end{align}
\begin{align}
    &i, j: \text{City index} \\
    &N: \text{Number of Cities} \\
    &c_{ij}: \text{Distance from city } i \text{ to city } j \\
    &x_{ij}: \begin{cases}
        1 & \text{if you move from city $i$ to city $j$}\\
        0 & \text{otherwise}
    \end{cases}.\\
    &u_i: \text{arbitrary numbers representing the order of city $i$ in the tour}
\end{align}
The constraints~\eqref{atsp_constraint1} and ~\eqref{atsp_constraint2} ensure that each city is visited exactly once. Constraint~\eqref{atsp_constraint3} prevents subtours, ensuring that all cities are included in a single tours of length $n$.

\paragraph{Heuristics}
As the name suggests, Nearest Neighbor (NN), Nearest Insertion (NI), and Furthest Insertion (FI) are straightforward simple greedy-selection algorithms frequently used as baselines for TSP algorithms. We use the implementations provided by ~\citep{kwon2021matrix} which are implemented in C++.

LKH3 is a widely recognized state-of-the-art algorithm for addressing constrained TSP and Vehicle Routing Problems (VRP). It employs a local search approach utilizing $k$-opt operations to enhance its solutions. For solving the ATSP instances, we utilize version 3.0.6.

\subsection{Training Detail}\label{atsp train detail}
\paragraph{MatNet}
We utilize the checkpoints provided by.~\citep{kwon2021matrix} and evaluate them on the same problem instances.
\paragraph{Diffusion (SL)}
Using LKH-3, we generate 128,000 training samples for supervised learning. We use DIFUSCO's encoder, employing 5 layers for encoder. We set the denoising timestep to 20. To enforce feasible solution, we apply greedy decoding to the heatmap generated after the denoising process. The greedy decoding starts at node 0 and chooses the maximum probability node among those not visited. All other hyperparameters, follow those reported in ~\citep{sun2023difusco}.
\paragraph{Diffusion (RL)}
We train a standard discrete diffusion model with the objective function  \\
$\nabla_\theta \mathcal{J}_{\text{DDRL}}=\mathbb{E}\left[ \sum_{t=0}^{T}\nabla_\theta\log p_\theta(X_{t-1}|X_t,c)r(X_0,c) \right]$ proposed by~\citep{black2023training}. We use the same encoder architecture as IC/DC, along with a GAT-based decoder, consisting of 5 encoder layers and 2 decoder layers. Additionally, we set the denoising timestep to 20. To enforce the model to output feasible solution, we apply greedy decoding, which is the same as the one used in Diffusion (SL). All other hyperparameters are consistent with those used in IC/DC.

\paragraph{IC/DC}
For ATSP-20 instances, we use 3 layers for both the encoder and the decoder, with the denoising timestep set to 10. For ATSP-50 instances, we utilize 5 encoder layers and 3 decoder layers, with the denoising timestep set to 15. The hyperparameters $\lambda_1$ and $\lambda_2$ are set to 1e-3 and 1e-6, respectively. The \texttt{IMPROVEMENT} step is executed every 30 epochs. Our model is trained using the Adam optimizer, with a batch size of 256 for 20 node instances and 64 for 50 node instances and a learning rate of 4e-4.

\subsection{Complex Feature Data}\label{app_real}
\begin{table*}[h] 
    \centering
    \begin{tabular}{llccc}
        \toprule
        & & \multicolumn{3}{c}{NP-20}\\
        \cmidrule(lr){3-5}
        & & Tour Time $\downarrow$ & Gap $\downarrow$ & Time $\downarrow$\\
        \midrule
        (Meta-)heuristics & Random & 3.50 & 76.43\% & (0s) \\
        \midrule
        Auto-regressive & MatNet ($\times 256$) & 2.60 & 49.95\% & (1m)  \\
        \midrule
        Ours & \textbf{IC/DC* ($\times 64$)} & \textbf{1.56} & 0\% & (1m) \\ 
        \bottomrule
    \end{tabular}
    \caption{\textbf{(Results for NP-20)}. * denotes the baseline for computing the performance gap.} 
    \label{table:NP result} 
\end{table*}

In real-world combinatorial optimization (CO) problems, direct problem data, such as the distance matrix in ATSP or the processing time matrix in PMSP, may not always be explicitly provided. In such cases, solving the problem may require processing a broader range of data. For example, consider a navigation problem (NP) similar to ATSP, where the objective is to minimize the total travel time rather than distance. In this scenario, various types of information that influence travel time must be considered, including the coordinates of each city, time-per-distance data representing the relationships between cities, and traffic information.

We define the actual travel time between $|\mathcal{N}|$ cities as follows:
\begin{equation}
[T]^a_{i,j} = ([R]_i - [R]_j)^2 \cdot [S]_{i,j} + [F]_{i,j}
\end{equation}
where, for each \(i,j \in \mathcal{N}\), \(T^a \in \mathbb{R}^{|\mathcal{N}| \times |\mathcal{N}|}_+\) represents the actual travel time matrix, \(R \in \mathbb{R}^{|\mathcal{N}| \times 2}_+\) is the coordinate matrix of all cities, \(S \in \mathbb{R}^{|\mathcal{N}| \times |\mathcal{N}|}_+\) is the reciprocal speed matrix, and \(F \in \mathbb{R}^{|\mathcal{N}| \times |\mathcal{N}|}\) is the traffic matrix. The goal is to minimize the total \([T]^a_{i,j}\) across the entire route.

In this case, instead of the direct problem data \(T^a\), we need to consider the complex problem instance \(c = (\mathcal{N}, R, S, F)\). Typically, methods such as solvers or heuristics require significant expert effort to process such data.

\begin{figure*}[t]
    \centering
    \begin{minipage}[b]{0.5\textwidth}
        \centering
        \includegraphics[width=\textwidth]{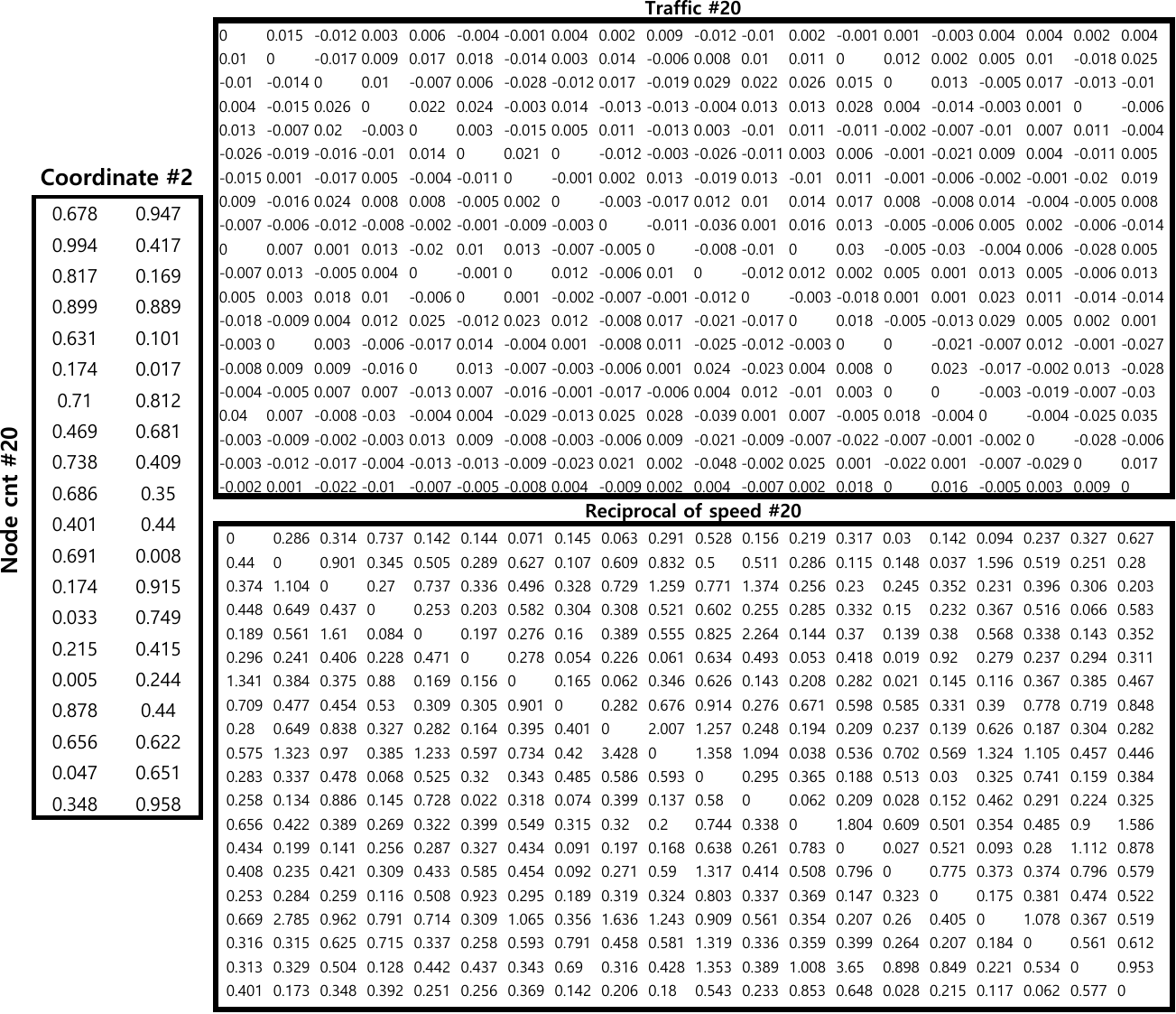}
        \caption{Problem Instance $c$}
    \end{minipage}
    \hspace{0.05\textwidth}
    \begin{minipage}[b]{0.4\textwidth}
        \centering
        \includegraphics[width=\textwidth]{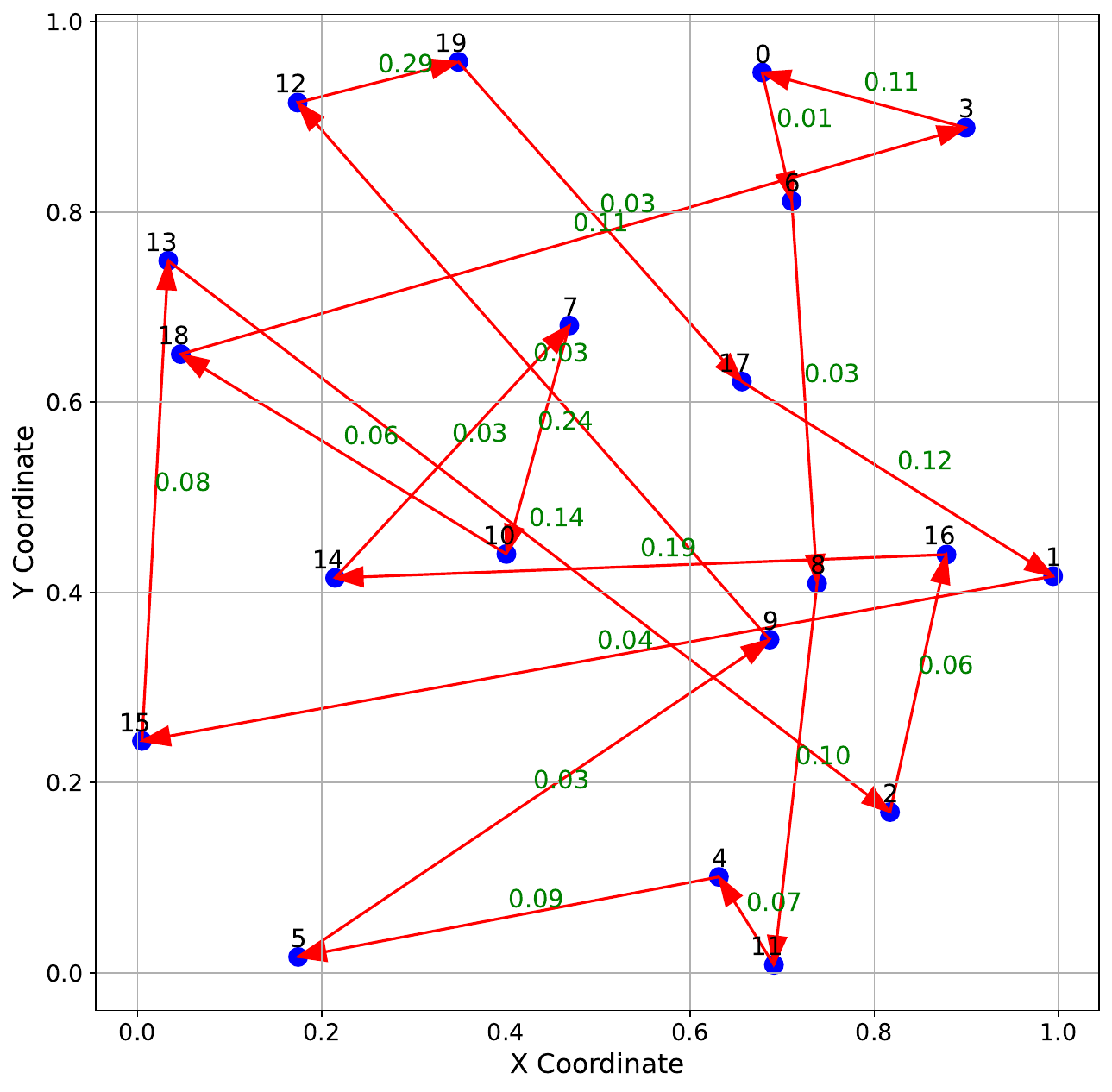}
        \caption{Solution Visualization}
    \end{minipage}
    \caption{\textbf{(Left)} The instance \(c\) is depicted with the coordinate matrix \(R\) on the left, the traffic matrix \(F\) on the top right, and the reciprocal of speed matrix \(S\) on the bottom right. \textbf{(Right)} The figure illustrates the solution for the instance shown on the left. The blue dots represent the coordinates of each city, while the red arrows indicate the travel direction between cities in the solution. The green numbers along the arrows represent the actual travel time between the connected cities.}
    \label{real_ATSP}
\end{figure*}

On the other hand, IC/DC is capable of handling these diverse types of data and demonstrates strong generalization performance. By the  problem encoder, the information \(R\) for each city is processed through an embedding layer with dimension \(d\) and then sent to \(A\) and \(B\). The relational information between cities, \(S\) and \(F\), is input as \(D = S||F \in \mathbb{R}^{|\mathcal{N}| \times |\mathcal{N}| \times 2}\). As shown in~\cref{table:NP result}, IC/DC outperforms other baselines and demonstrates strong potential for generalizability. Additionally, an example of the NP is illustrated in~\cref{real_ATSP}.

\section{Additional experiment results}\label{additional_experiments}
\subsection{Distribution generalization full result}\label{generalizability_full_result}
\begin{table}[h]
    \centering
    \setlength{\tabcolsep}{5pt} 
    \begin{tabular}{ccccccccc}
        \hline
         & \multicolumn{2}{c}{MatNet} & \multicolumn{2}{c}{Diffusion (SL)} & \multicolumn{2}{c}{Diffusion (RL)} & \multicolumn{2}{c}{IC/DC} \\
        \cmidrule(lr){2-3}\cmidrule(lr){4-5}\cmidrule(lr){6-7} \cmidrule(lr){8-9}
        Node & Tour Length $\downarrow$ & Gap $\downarrow$ & Tour Length $\downarrow$ & Gap $\downarrow$ & Tour Length $\downarrow$ & Gap $\downarrow$ & Tour Length $\downarrow$ & Gap $\downarrow$ \\
        \hline
        10 & 1.498 & 0.064\% & 1.744 & 15.213\% & 2.403 & 46.427\% & 1.497 & \textbf{0}\%\\
        15 & 1.590 & 4.009\% & 1.765 & 14.411\% & 2.815 & 59.275\% & 1.528 & \textbf{0}\% \\
        20* & 1.534 & 0.013\%  & 1.755 & 13.438\% & 3.331 & 73.875\% & 1.534 & \textbf{0}\% \\
        25 & 2.006 & 25.712\% & 1.934 & 22.147\% & 3.328 & 72.986\% & 1.553 & \textbf{0.255}\% \\
        30 & 2.243 & 36.678\% & 2.069 & 28.803\% & 3.501 & 77.375\% & 1.607 & \textbf{3.733}\% \\
        35 & 2.462 & 45.888\% & 2.213 & 35.647\% & 3.678 & 81.762\% & 1.765 & \textbf{13.393}\% \\
        40 & 2.640 & 51.849\% & 2.389 & 42.407\% & 3.800 & 83.971\% & 1.976 & \textbf{23.987}\% \\
        45 & 2.798 & 57.116\% & 2.533 & 47.843\% & 3.936 & 86.743\% & 2.174 & \textbf{33.227}\% \\
        50 & 2.950 & 62.150\% & 2.676 & 53.233\% & 4.090 & 90.010\% & 2.421 & \textbf{43.797}\% \\
        \hline
    \end{tabular}
    \caption{Full result on distribution generalization evaluation across different methods. * denotes the training distribution (20 nodes).}
    \label{tab:generalizability_full}
\end{table}

~\cref{tab:generalizability_full} shows the generalization performance of various models, including MatNet, Diffusion (SL), Diffusion (RL), and IC/DC, evaluated on node sizes from 10 to 50. Notably, IC/DC, trained on 20-node distributions, achieves a perfect gap of 0\% at its training distribution and smaller node sizes (10 and 15 nodes), outperforming all other models.

As node size increase beyond the training distribution, IC/DC continues to exhibit strong generalization capabilities. At 30 nodes, IC/DC achieves a gap of 3.733\%, outperforming MatNet (36.678\%) and Diffusion models (SL: 28.803\%, RL: 77.375\%). Even at 50 nodes, IC/DC maintains its superiority with a gap of 43.797\%, significantly better than MatNet (62.150\%) and Diffusion models (SL: 53.233\%, RL: 90.010\%).

In summary, IC/DC consistently delivers superior results across various node sizes, demonstrating excellent adaptability and generalization. Its ability to maintain competitive performance, even on test distributions far from its training distribution, highlights its efficiency and robustness in CO problems.


\begin{figure*}[t]
    \centering
    \includegraphics[width=\textwidth]{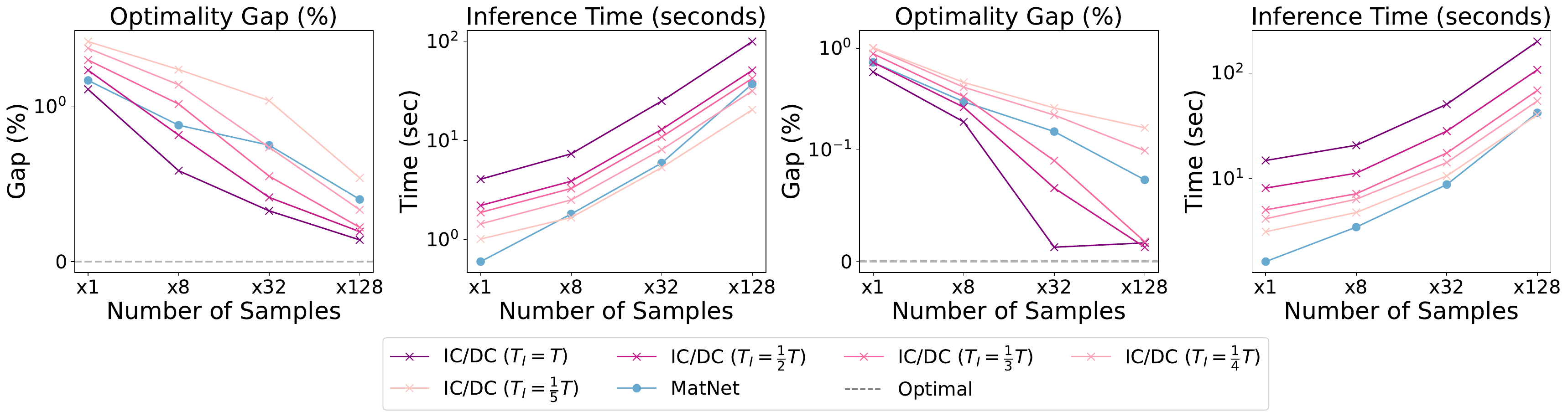}
    \caption{Full results on optimality gap and inference time as a function of sampling step count in the parallel machine scheduling problem (PMSP). The left two plots correspond to a 3 machines \& 50 jobs setup, while the right two plots represent a 5 machines \& 50 jobs setup.}
    \label{fast_inference}
\end{figure*}

\subsection{Reducing sampling steps for faster inference additional results}
\label{Reducing_sampling_faster_inference_additional_results}
\cref{fast_inference} demonstrates the relationship between varying sampling steps and their impact on the optimality gap and inference time. The results are shown for IC/DC and MatNet when solving PMSP instances with 20 jobs on 5 machines and 50 jobs on 3 machines. While neither IC/DC nor MatNet achieves a 0\% optimality gap, IC/DC performs better in terms of optimality gap compared to MatNet at $T_I=T$.

The larger the number of samples during inference, the less the optimality gap is affected when reducing sampling steps ($T_I$). IC/DC outperforms MatNet at $T_I=\frac{1}{3}T$ and $T_I=\frac{1}{4}T$. Moreover, IC/DC achieves superior performance while maintaining competitive inference speed.

\end{document}